\documentclass[journal]{IEEEtran}
\IEEEoverridecommandlockouts
\usepackage{cite}
\usepackage{amsmath,amssymb,amsfonts}
\usepackage{algorithmic}
\usepackage{graphicx}
\usepackage{textcomp}
\usepackage{xcolor}
\usepackage{hyperref}
\usepackage{subcaption}
\usepackage{xcolor}
\usepackage{algorithm}
\hypersetup{hidelinks} 
\usepackage{caption}
\def\BibTeX{{\rm B\kern-.05em{\sc i\kern-.025em b}\kern-.08em
    T\kern-.1667em\lower.7ex\hbox{E}\kern-.125emX}}
\begin{document}

\title{\fontsize{22.2}{28}\selectfont FedPGT: Progressive Gradient Transmission for Vehicular Federated Learning over Time-Varying Channels
}
\author{\IEEEauthorblockN{Jintao Yan,~\IEEEmembership{Graduate Student Member,~IEEE,} Tan Chen, Yuxuan Sun,~\IEEEmembership{Member,~IEEE},\\  Sheng Zhou,~\IEEEmembership{Senior Member,~IEEE} and Zhisheng Niu,~\IEEEmembership{Fellow,~IEEE} \vspace{-2.5em}}\\
\thanks{J. Yan, T. Chen, S. Zhou (Corresponding Author) and Z. Niu are with the Beijing National Research Center for Information Science and Technology, Department of Electronic Engineering, Tsinghua University, Beijing 100084, China. (email: \{yanjt22, chent21\}@mails.tsinghua.edu.cn, \{sheng.zhou, niuzhs\}@tsinghua.edu.cn). 

Y. Sun is with the School of Electronic and Information Engineering, Beijing Jiaotong University, Beijing 100044, China. (e-mail: yxsun@bjtu.edu.cn).}
}

\maketitle

\maketitle


\begin{abstract}
Vehicular federated learning (VFL) enables privacy-preserving collaborative model training for intelligent transportation systems, where communication resource allocation and gradient sparsification techniques have been explored to reduce communication overhead. However, vehicle mobility leads to rapidly varying channel conditions and transmission capacity, rendering predetermined resource allocation and sparsification decisions ineffective. In this paper, we propose FedPGT, a progressive gradient transmission scheme for VFL over time-varying channels, where vehicles progressively transmit high-magnitude gradient entries in response to instantaneous channel conditions. We establish a convergence bound that characterizes the impact of transmitted gradient entries and reveals diminishing-return behavior governed by a power-law decay. Motivated by this result, we formulate a stochastic optimization problem for online decision-making, where the main challenge lies in a cumulatively coupled, non-separable objective. To handle this challenge, we introduce per-slot surrogate transmission variables to decouple the long-term dependence across time slots and convert the original objective into an additive per-slot optimization problem, enabling a Lyapunov drift-plus-penalty approach for online scheduling. We further develop a low-complexity resource allocation algorithm for efficient online implementation. Experimental results demonstrate that the proposed scheme achieves a $3.65\%$ accuracy improvement on the CIFAR-10 image classification task and a $12.66\%$ reduction in average displacement error on the Argoverse trajectory prediction task compared with state-of-the-art baselines, demonstrating its applicability to diverse learning tasks under highly dynamic vehicular environments.

\end{abstract}
\section{Introduction}

The rapid development of vehicular networks has enabled a wide range of data-driven applications, such as cooperative perception, trajectory prediction, and intelligent route planning \cite{sun2020edge}. These applications rely on machine learning (ML) models that require continuous adaptation to cope with dynamic traffic conditions, evolving environments, and diverse driving behaviors. Traditional centralized learning frameworks require raw data to be uploaded to a central server for model training \cite{FLreview, Xu2023EdgeLearning, Jia2025Survey}, which not only incurs excessive communication latency but also raises serious privacy concerns, particularly when sensitive sensor data are involved.

With the increasing deployment of on-board sensing and computing capabilities, vehicles are becoming capable of performing local model training using their own data. This trend has driven the adoption of federated learning in vehicular networks, where multiple vehicles collaboratively train a global model by exchanging model updates instead of raw data. Vehicular federated learning (VFL) enables timely model adaptation while preserving data privacy and has emerged as a promising paradigm for large-scale intelligent transportation systems \cite{VFLS2, yan}.  

In a typical VFL process, each participating vehicle performs local training using its private data and periodically uploads model updates (e.g., gradients or parameter differences) to a roadside unit (RSU) or aggregation server. The server aggregates the received updates and broadcasts the updated global model to the vehicles for the next training round. This process repeats until convergence.

Despite its advantages, VFL faces significant challenges due to the highly dynamic nature of vehicular networks \cite{VFL8, Yan2025DynamicSchedulingV2VFL}. First, vehicles remain within the coverage area of an RSU only for a limited sojourn time, which constrains the time available for model transmission. Second, wireless channel conditions fluctuate rapidly because of vehicle mobility, leading to time-varying uplink capacities and intermittent communication opportunities. As a result, reliable and efficient model transmission becomes a major bottleneck in VFL. 

To alleviate the communication overhead, many existing works adopt model compression or gradient sparsification techniques, which reduce the number of transmitted parameters by retaining only the most informative components. In these approaches, the sparsification degree, which is defined as the number of gradient entries transmitted in each round, is typically fixed at the beginning of the model uploading \cite{Li2021ToTalkOrWork, Lin2023JGSD}. However, such a design is not well-suited for vehicular environments, where the achievable number of transmitted gradient entries during a round is highly uncertain due to rapidly varying channels and dynamic communication opportunities. Moreover, most existing sparsification-based convergence analyses assume a fixed set of participating devices \cite{Hu2024HybridGradientCompression}, which does not hold in vehicular networks where the set of participating vehicles changes across training rounds. These limitations suggest that the number of transmitted gradient entries should adapt to the dynamically changing communication opportunities in vehicular networks, rather than being predetermined before transmission.

In this work, we propose FedPGT, a progressive gradient transmission scheme for VFL that adapts to time-varying communication conditions and dynamic vehicle participation. Instead of predefining the sparsification degree, vehicles progressively transmit high-magnitude gradient entries based on instantaneous communication conditions. The main contributions are summarized as follows:

\begin{itemize}
\item We establish a convergence analysis that characterizes the relationship between the number of transmitted gradient entries and the convergence performance of VFL, showing that increasing the number of transmitted gradient entries improves convergence performance with diminishing returns following a power-law decay. This result indicates that transmitting only a small number of high-magnitude gradient entries can preserve most of the convergence gain, which motivates progressive gradient transmission under limited communication opportunities.

\item We propose a progressive gradient transmission scheme for VFL. We formulate a stochastic optimization problem, where the main challenge lies in a cumulatively coupled, non-separable objective. To handle this challenge, we introduce per-slot surrogate transmission variables to decouple the long-term dependence across time slots and construct an upper-bounding reformulation that converts the original problem into an additively separable per-slot optimization problem. Based on this transformation, we develop a Lyapunov drift-plus-penalty approach for online scheduling and transmission decisions without requiring future channel information.

\item We derive structural properties for the online resource allocation problem based on utility-based optimality conditions. In particular, we show that exclusive resource block (RB) allocation remains optimal despite relaxation and obtain an analytical power allocation expression. Based on these results, we develop a low-complexity online resource allocation algorithm for efficient implementation in dynamic vehicular environments.

\item Experimental results show that, compared with state-of-the-art benchmarks, the proposed scheme improves the test accuracy by $3.65\%$ on the CIFAR-10 image classification task and reduces the average displacement error (ADE) by $12.66\%$ on the Argoverse trajectory prediction task, demonstrating the effectiveness and applicability of FedPGT across diverse learning tasks.

\end{itemize}

The remainder of this paper is organized as follows. Section II presents a review of related works. Section III describes the system model. Section IV develops the convergence analysis and formulates the optimization problem. Section V introduces the proposed progressive gradient transmission scheme. The experimental evaluations are presented in Section VI, followed by the conclusions in Section VII.

\begin{figure*}[t!]
\centering
\includegraphics[width=0.8\textwidth]{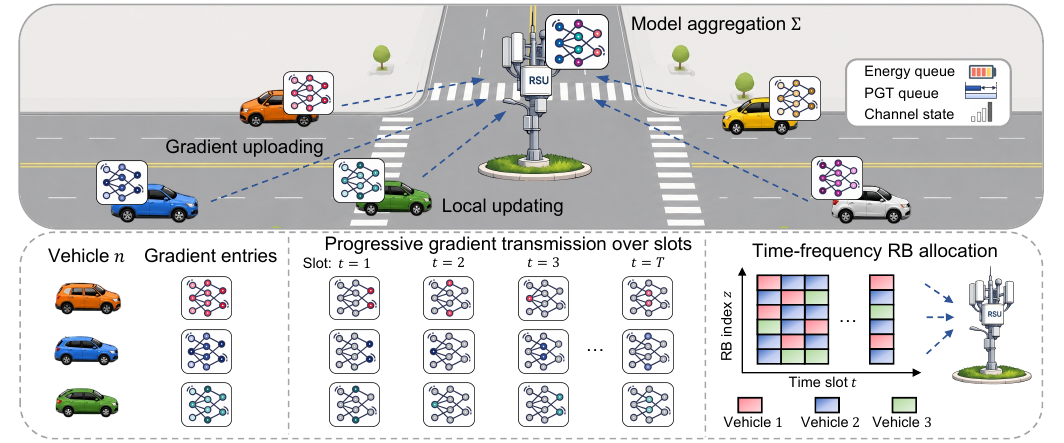}
\caption{The proposed FedPGT scheme.}
\label{system}
\end{figure*}

\section{Related Work}
\subsection{Federated Learning in Vehicular Networks}

Federated learning over wireless networks has attracted significant attention due to the tight coupling between learning performance and communication constraints. Comprehensive surveys such as \cite{FLreview, Xu2023EdgeLearning, Jia2025Survey} summarize recent advances in federated learning in wireless systems, highlighting challenges arising from limited bandwidth \cite{DS6, DS7}, fading channels \cite{DS10, DSsun, DS9}, and device heterogeneity \cite{chentan}.

Motivated by intelligent transportation applications, federated learning has been extended to vehicular networks, where high mobility and intermittent connectivity pose additional challenges. The authors of \cite{VFL6, VFL7, VFLnew1, VFLnew6} study vehicle selection and resource optimization for vehicular edge FL, considering the limited sojourn time of vehicles. To enhance participation and robustness in dynamic environments, incentive mechanisms have been proposed to motivate vehicle collaboration in federated learning \cite{VFLnew5, RLIncentiveVFL}. In \cite{MobilityAwareMTVFL}, mobility-aware decentralized federated learning frameworks have been developed to explicitly model vehicle mobility and multi-task learning dynamics in vehicular networks. Semi-asynchronous federated learning schemes have also been proposed to improve robustness against mobility-induced communication dynamics \cite{VFLnew2}. In addition, split federated learning and semantic communication enhanced frameworks have been investigated to improve communication efficiency in resource-constrained vehicular networks \cite{VFLnew3, VFLnew4}. These works demonstrate the potential of FL in vehicular systems but typically operate at the level of training rounds.

More recently, researchers have begun to investigate slot-level communication and resource optimization in vehicular FL systems. The authors of \cite{VFL8} consider resource-constrained vehicular edge FL with highly mobile vehicles and analyze the impact of mobility on learning performance. The authors of \cite{Yan2025DynamicSchedulingV2VFL} develop a dynamic scheduling scheme for vehicle-to-vehicle communications enhanced FL by adapting resource allocation to time-varying channels. These works explicitly consider slot-level communication dynamics and online resource allocation. However, the structure of model update transmission is still typically predetermined before transmission. They do not explicitly model the progressive transmission of dominant gradient entries within each round under rapidly varying communication conditions, which is the key problem addressed in this work.

\subsection{Compression and Sparsification in Federated Learning}

Communication-efficient learning has been widely studied through gradient and model compression techniques, among which sparsification transmits only a subset of gradient entries to reduce communication overhead. Early theoretical works typically assume a fixed number of transmitted gradient entries per round, such as top-$k$ or random-$k$ sparsification. A representative example is \cite{Stich2018SparsifiedSGDMemory}, which establishes convergence guarantees for $k$-sparsified SGD under standard assumptions. Qsparse-local-SGD \cite{Basu2020QsparseLocalSGD} further combines sparsification with quantization and local updates, providing convergence results for both convex and non-convex objectives. More recent works investigate adaptive compression mechanisms that adjust the number of transmitted gradient entries across training rounds according to training dynamics or data heterogeneity. For instance, the authors of \cite{Han2020AdaptiveGradientSparsification} formulate adaptive gradient sparsification in federated learning as an online decision problem to balance communication and computation. Hybrid compression schemes that integrate sparsification with quantization or encoding have also been explored to improve efficiency \cite{Hu2024HybridGradientCompression}. Nevertheless, most of these approaches determine the sparsification degree at the beginning of each training round and assume a fixed communication budget within the round.

Beyond compression-only designs, several studies jointly optimize sparsification and communication scheduling. The authors of \cite{Li2021ToTalkOrWork} propose a flexible compression control mechanism for mobile edge FL to balance local computation and wireless transmission under energy constraints. The authors of \cite{Lin2023JGSD} investigate joint gradient sparsification and device scheduling under limited communication resources. The authors of \cite{Wei2024GradientSparsificationWireless} further integrate sparsification with wireless resource allocation under privacy constraints. However, most existing joint designs assume that the sparsification structure or degree is decided \emph{a priori} per round (or changes only at round boundaries), and they typically do not model the \emph{progressive, within-round} transmission of dominant gradient entries under fast channel fluctuations, which is a key feature in vehicular networks considered in this work.

\section{System Model}
\label{II}
\subsection{VFL Model}

We study a VFL system, as depicted in Fig. \ref{system}, where an RSU coordinates the collaborative model training among vehicles within its coverage area through $R$ rounds of training. During the $r^{\text{th}}$ training round, the set of vehicles participating in the model training are denoted by $\mathcal{N}^{(r)}=\{1,2,\ldots,N^{(r)}\}$. Each vehicle $n\in \mathcal{N}^{(r)}$ holds a local dataset drawn from distribution $\mathcal{X}_n$ over the input space $\mathcal{D}_n$. For a given sample $\boldsymbol{d} \in \mathcal{D}_n$, the loss function $f(\boldsymbol{w}, \boldsymbol{d})$ quantifies how well the model $\boldsymbol{w}$ fits the data. Accordingly, the local loss function for vehicle $n$ is defined as the expected loss over its data distribution:
$$F_n(\boldsymbol{w})\triangleq \underset{\boldsymbol{d}\sim\mathcal{X}_n}{\operatorname*{\mathbb{E}}}[f(\boldsymbol{w},\boldsymbol{d})].$$
Unlike conventional FL settings with a static client set, the participating vehicles in VFL vary over time due to mobility. We assume that vehicles participating in each round are sampled from an underlying distribution $\mathcal{P}$. The global loss function is therefore defined as
\begin{equation} F(\boldsymbol{w}) \triangleq \underset{n\sim\mathcal{P}}{\operatorname*{\mathbb{E}}}[F_n(\boldsymbol{w})]. \label{global} \end{equation}
The objective is to minimize the global loss function in $R$ communication rounds by optimizing the model parameters $\boldsymbol{w}$. The index set of communication rounds is denoted by $\mathcal{R} = \{1,2,\ldots,R\}$.
The VFL training procedure in each round comprises three key stages: local updates, gradient uploading, and model aggregation.

\subsubsection{Local Updates} At the beginning of round $r$, the RSU broadcasts the current global model $\boldsymbol{w}^{(r-1)}$ to all vehicles within its coverage. Upon reception, each vehicle $n \in \mathcal{N}^{(r)}$ computes its stochastic gradient based on stochastic gradient descent (SGD):
\begin{equation}\boldsymbol{g}_n^{(r)} = \frac{1}{|\mathcal{B}^{(r)}_n|}\sum_{\boldsymbol{d}\in\mathcal{B}^{(r)}_n} \nabla f\left(\boldsymbol{w}^{(r-1)},\boldsymbol{d}\right),\label{GD}\end{equation}
where $\mathcal{B}_{n}^{(r)} \subseteq \mathcal{D}_n$ denotes a mini-batch sampled from $\mathcal{D}_n$.

\subsubsection{Gradient Uploading} After local training, vehicles progressively transmit dominant gradient entries to the RSU according to the proposed transmission strategy. Let $\Psi_n^{(r)}(\cdot)$ denote the resulting gradient transmission operator for vehicle $n$ in round $r$. The communication mechanism supporting this stage is described in Section~\ref{comm}, while the detailed definition of $\Psi_n^{(r)}(\cdot)$ is provided in Section~\ref{secspar}.

\subsubsection{Model Aggregation} After receiving the uploaded gradients, the RSU aggregates them to update the global model:
\begin{equation}\boldsymbol{w}^{(r)}=\boldsymbol{w}^{(r-1)} - \frac{\eta}{N^{(r)}}\sum_{n\in \mathcal{N}^{(r)}} \Psi_n^{(r)}(\boldsymbol{g}_n^{(r)}),
\label{agg1}\end{equation}
and proceeds to the next training round.
\subsection{Computation Model}
We adopt a standard computation model \cite{comp1, comp2} for local model updates. Let $N_{\text{flop}}$ denote the number of floating-point operations (FLOPs) required to process one training sample. For vehicle $n$ in round $r$, the CPU clock frequency is denoted by $l_n^{(r)}$ (cycles/s). Accordingly, the computation latency for local training is given by
\begin{equation}
\rho_{n}^{(r)} = \frac{N_{\text{flop}} |\mathcal{B}^{(r)}_n|}{l_{n}^{(r)}},\notag 
\end{equation}
while the corresponding computation energy consumption is
\begin{equation}
\xi_{n}^{(r)} = \theta (l_{n}^{(r)})^2N_{\text{flop}} |\mathcal{B}^{(r)}_n|,\notag 
\end{equation}
where $\theta$ denotes the effective switched capacitance coefficient determined by the processor chip architecture.
\subsection{Communication Model}
\label{comm}
We consider an OFDMA-based vehicular network operating over discrete time slots. For round $r$, the set of time slots is defined as $\mathcal{T}^{(r)}=\{1,2,\ldots,T\}$, where $T$ denotes the number of slots per round and $\tau$ represents the duration of each slot.

For downlink model distribution, the RSU broadcasts the global model to all vehicles using the entire bandwidth and sufficiently high transmission power. Therefore, following \cite{VFL8, Yan2025DynamicSchedulingV2VFL}, the downlink communication latency is ignored. For uplink gradient transmission, we consider a single-input-multiple-output (SIMO) system, where each vehicle is equipped with a single antenna and the RSU is equipped with $M$ antennas. The uplink bandwidth is equally divided into $Z$ orthogonal resource blocks (RBs), each with bandwidth $\beta$. In each slot, the RSU allocates the RBs to vehicles, and the vehicles upload their gradients using the allocated RBs. We use $s_{n,z}(t)$ to denote the RB allocation indicator, where $s_{n,z}(t)=1$ if RB $z$ is allocated to the vehicle $n$ in slot $t$. Since one RB can only be allocated to one vehicle, $s_{n,z}(t)$ has the following constraints:
\begin{equation}
\sum_{n \in \mathcal{N}^{(r)}} s_{n,z}(t) \leq 1, \quad \forall z \in \mathcal{Z}, \ \forall t \in \mathcal{T}^{(r)}, \label{trans1}
\end{equation}
\begin{equation}
s_{n,z}(t) \in \{0,1\}, \quad \forall n \in \mathcal{N}^{(r)}, \forall z \in \mathcal{Z}, \ \forall t \in \mathcal{T}^{(r)}. \label{trans2}
\end{equation}
We denote the transmission power allocated to RB $z$ by $p_{n,z}(t)$, which is constrained by
\begin{equation}0\le \sum_{z\in \mathcal{Z}} p_{n,z}(t)\le p^{\text{max}}_n,\quad \forall n \in \mathcal{N}^{(r)},\ \forall t \in \mathcal{T}^{(r)}. \label{power1}\end{equation}
We denote the transmitted signal of vehicle $n$ in slot $t$ by $x_n(t)$. The received signal at the RSU over RB $z$ is given by
$$y_{n,z}(t) = s_{n,z}(t) \sqrt{p_{n,z}(t)} \boldsymbol{u}_{n,z}(t)^H \boldsymbol{h}_{n,z}(t)x_n(t)+n_0,$$
where $\boldsymbol{u}_{n,z}(t)\in\mathbb{C}^{M\times1}$ is the receiver beamforming vector. $n_0\sim \mathcal{CN}(0,N_0)$ denotes additive white Gaussian noise. The channel vector $\boldsymbol{h}_{n,z}(t)\in\mathbb{C}^{M\times1}$ is modeled as
$$\boldsymbol{h}_{n,z}(t) = \sqrt{\psi_{n}(t)} \varrho_{n}(t)\hat{\boldsymbol{h}}_{n,z}(t),$$ where $\sqrt{\psi_{n}(t)}$ denotes the large-scale path loss, $\varrho_{n}(t)$ denotes the shadowing fading coefficient, and $\hat{\boldsymbol{h}}_{n,z}(t)$ denotes the small-scale fading vector. 

The RSU uses maximal-ratio combining receiver beamforming, i.e., $\boldsymbol{u}_{n,z}(t) = \boldsymbol{h}_{n,z}(t)/\left\Vert \boldsymbol{h}_{n,z}(t) \right\Vert$. Then, the received signal-to-noise ratio (SNR) at the RSU from vehicle $n$ over RB $z$ in slot $t$ is given by $$\Gamma_{n,z}(t) = \frac{ p_{n,z}(t) \big\Vert \boldsymbol{h}_{n,z}(t) \big\Vert^2}{\beta N_0},$$

Based on the above model, the uplink transmission rate of vehicle $n$ in slot $t$ is given by 
\begin{equation}
A_n(t) = \sum_{z\in\mathcal{Z}} s_{n,z}(t)\beta \log_2 (1+ \Gamma_{n,z}(t)),
\end{equation}
while the corresponding transmission energy consumption is
\begin{equation}
e_n(t) = \sum_{z\in\mathcal{Z}} \tau s_{n,z}(t) p_{n,z}(t).
\end{equation}

\subsection{Progressive Gradient Transmission}
\label{secspar}

Based on the above communication model, vehicles progressively upload dominant gradient entries over the slots $t\in\mathcal{T}^{(r)}$ according to the available communication opportunities.

The communication overhead of each transmitted gradient entry consists of $\lceil\log_2 I\rceil$ bits for the index and $b$ bits for the value, where $I$ denotes the gradient dimension and $b=32$ corresponds to single-precision floating-point representation. The number of new gradient entries that vehicle $n$ can upload in slot $t$ is integer-valued and capped by
\begin{equation}
\kappa_n(t)
=
\min\left\{
\left\lfloor
\frac{\tau A_n(t)}
{b+\lceil\log_2 I\rceil}
\right\rfloor,
I-\sum_{t'=1}^{t-1}\kappa_n(t')
\right\},
\label{sparrate}
\end{equation}
where $\tau$ denotes the slot duration and $A_n(t)$ denotes the achievable uplink rate in slot $t$. Accordingly, the total number of uploaded gradient entries in round $r$ is $k_n^{(r)}
=
\sum_{t\in\mathcal{T}^{(r)}}
\kappa_n(t)$.
For analytical tractability, the resource optimization in Sections~V-B and V-C uses the continuous relaxation
\begin{equation}
\bar{\kappa}_n(t)
=
\frac{\tau A_n(t)}
{b+\lceil\log_2 I\rceil},
\label{sparrate_relax}
\end{equation}
To avoid allocating RBs or power to vehicles that have already uploaded all gradient entries, we impose
\begin{equation}
\begin{aligned}
&s_{n,z}(t)=0,\quad p_{n,z}(t)=0,\\
&\forall n\in\mathcal N^{(r)},\ \forall z\in\mathcal Z,\ \forall t\in\mathcal T^{(r)},\ \sum_{t'=1}^{t-1}\kappa_n(t')\ge I.
\end{aligned}
\label{fullupload}
\end{equation}
After the RB and power allocation decisions are obtained, the actual number of transmitted entries is computed by the integer rounding and capping rule in \eqref{sparrate}. The continuous relaxation together with \eqref{fullupload} is not exactly equivalent to \eqref{sparrate}. However, since the model dimension $I$ is typically very large, the resulting approximation error is negligible.

The gradient transmission operator $\Psi_n^{(r)}(\boldsymbol g_n^{(r)})$ is then formally defined as
\begin{equation}
\big[\Psi_n^{(r)}(\boldsymbol g_n^{(r)})\big]_i
=
\begin{cases}
[\boldsymbol g_n^{(r)}]_i,
& i \in \mathcal K_n^{(r)},
\\
0,
& \text{otherwise},
\end{cases}
\end{equation}
where $\mathcal K_n^{(r)}$ denotes the index set corresponding to the $k_n^{(r)}$ largest-magnitude entries in $\boldsymbol g_n^{(r)}$.



\section{Problem Formulation}
\subsection{Convergence Analysis}

The goal of VFL is to minimize the global loss function (\ref{global}). However, the impact of the number of transmitted gradient entries on the global loss is implicit. Therefore, we derive a convergence bound to characterize the relationship between gradient transmission and learning performance. We first introduce the definition of compressible gradients \cite{compress0, compress1, compress2, compress3, compress4}.

\noindent \textbf{Definition 1 (Compressible Gradients).}
A vector $\boldsymbol{g}\in\mathbb{R}^{I}$ is said to be \emph{compressible} if the magnitudes of its entries, sorted in descending order, follow a power-law decay. Let $g_i$ denote the $i$-th largest-magnitude entry of $\boldsymbol g$. The vector $\boldsymbol g$ is compressible if there exist constants $C>0$ and $\alpha>\tfrac{1}{2}$ such that
\begin{equation}
|g_i| \le C i^{-\alpha},
\quad
\forall i\in\{1,2,\ldots,I\},
\end{equation}
where the compressibility exponent $\alpha$ characterizes the decay rate of the sorted entries, and a larger $\alpha$ indicates stronger compressibility. The scaling coefficient $C$ characterizes the overall magnitude of the vector.

Prior works have shown that stochastic gradients in deep learning are typically compressible \cite{compress0, compress1, compress2}. Accordingly, we make the following assumptions \cite{DS9, DS10,DSsun, CS4, CS5, Yan2025DynamicSchedulingV2VFL}.

\noindent \emph{Assumption 1:}
The stochastic gradient $\boldsymbol g_n^{(r)}\in\mathbb R^I$ generated by vehicle $n$ in round $r$ is compressible, i.e., there exist parameters $C_n^{(r)}>0$ and $\alpha_n^{(r)}>\tfrac{1}{2}$ such that
$$|g_{n,i}^{(r)}|
\le C_n^{(r)} i^{-\alpha_n^{(r)}},
\quad
\forall i\in\{1,2,\ldots,I\},$$
where $C_n^{(r)}$ and $\alpha_n^{(r)}$ can be estimated locally after each round by fitting the sorted gradient magnitudes to the power-law model.

\noindent \emph{Assumption 2:}
The variance of the stochastic gradient is bounded, i.e.,
$\mathbb{E}
\left\|
\boldsymbol g_n^{(r)}
-
\nabla F_n(\boldsymbol w^{(r-1)})
\right\|^2
\le
\sigma^2$,
where the expectation is taken over the randomness of SGD.

\noindent \emph{Assumption 3:}
The data heterogeneity induced by vehicle sampling is bounded, i.e., $\underset{n\sim\mathcal P}{\mathbb E}
\left\|\nabla F_n(\boldsymbol w)-\nabla F(\boldsymbol w)\right\|^2
\le \delta^2$ for any model parameter $\boldsymbol w$.

\noindent \emph{Assumption 4:}
The local loss function $F_n(\cdot)$ is $L$-smooth, i.e.,
\begin{equation}
\begin{aligned}
F_n(\boldsymbol w')
\le
F_n(\boldsymbol w)
+
\left<
\nabla F_n(\boldsymbol w),
\boldsymbol w'-\boldsymbol w
\right>
+
\frac L2
\left\|
\boldsymbol w'-\boldsymbol w
\right\|^2.
\notag
\end{aligned}
\end{equation}

For a given gradient $\boldsymbol g_n^{(r)}\in\mathbb R^I$, the gradient approximation error is defined as $\left\|
\boldsymbol g_n^{(r)}
-
\Psi_n^{(r)}(\boldsymbol g_n^{(r)})
\right\|^2$.

The following lemma characterizes the gradient approximation error under the compressible gradient assumption.

\noindent \textbf{Lemma 1 (Gradient Approximation Error Bound).}
The gradient approximation error of $\boldsymbol g_n^{(r)}$ is upper-bounded by
\begin{equation}
\left\|
\boldsymbol g_n^{(r)}
-
\Psi_n^{(r)}(\boldsymbol g_n^{(r)})
\right\|^2
\le
\frac{
2(C_n^{(r)})^2\alpha_n^{(r)}
(k_n^{(r)}+1)^{-(2\alpha_n^{(r)}-1)}
}{
2\alpha_n^{(r)}-1
}.
\end{equation}

\noindent \emph{Proof:} See Appendix \ref{Lemma1}. \hfill$\square$

Lemma~1 characterizes the approximation error caused by transmitting only partial gradient entries as a function of the number of uploaded gradient entries $k_n^{(r)}$. Building on Lemma~1, we derive the convergence bound.

\noindent \textbf{Theorem 1 (Convergence Bound). }  \emph{After $R$ rounds of training, the expected squared gradient norm of the global loss function is upper-bounded by}
\begin{equation}
\begin{aligned}
&\frac{1}{R}\sum_{r=0}^{R-1} \mathbb{E}\left\|\nabla F(\boldsymbol {w}^{(r)})\right\|^2 \\& \leq \frac{2\left(F(\boldsymbol{w}^{(0)})-\mathbb{E}[F(\boldsymbol{w}^{(R)})]\right)}{\eta R}+\frac{1}{R}\sum_{r=1}^{R}\frac{2(\sigma^2+\delta^2)}{N^{(r)}}\label{convbound}\\
& +\frac{1}{R}\sum_{r=1}^{R} \sum_{n\in \mathcal{N}^{(r)}} \frac{4(C_n^{(r)})^2\alpha_n^{(r)} (k_n^{(r)}+1)^{-(2\alpha_n^{(r)}-1)}}{N^{(r)}(2\alpha_n^{(r)}-1)}.  
\end{aligned}
\end{equation}
\emph{Proof:} See Appendix \ref{Theorem1}. \hfill$\square$ 

\noindent\textbf{Remark 1.} Theorem~1 shows that the convergence performance improves as the number of uploaded gradient entries $k_n^{(r)}$ increases. However, the improvement exhibits diminishing marginal returns. Specifically, due to the power-law term $(k_n^{(r)}+1)^{-(2\alpha_n^{(r)}-1)}$, each additional transmitted entry yields progressively smaller reductions in the gradient approximation error, and therefore smaller improvements in convergence performance. Consequently, transmitting a small set of dominant gradient entries captures most of the benefit, while transmitting additional entries provides limited further improvement.


\subsection{Problem Formulation}
Based on Theorem 1, we minimize the convergence upper bound in \eqref{convbound}. This is equivalent to minimizing $\sum_{n\in \mathcal{N}^{(r)}} \frac{(C_n^{(r)})^2\alpha_n^{(r)} (k_n^{(r)}+1)^{-(2\alpha_n^{(r)}-1)}}{2\alpha_n^{(r)}-1}$ in each round since other parameters are constant. The proposed scheme optimizes the communication resource allocation to indirectly control the progressive gradient transmission process. The optimization problem is formulated as
\begin{subequations}
\begin{align}
P0:  &\underset{\boldsymbol{S}^{(r)},\boldsymbol{P}^{(r)}}{\min}  \sum_{n\in\mathcal{N}^{(r)}} \frac{(C_n^{(r)})^2 \alpha_n^{(r)}}{2 \alpha_n^{(r)}-1} \left(\sum_{t\in\mathcal{T}^{(r)}}\kappa_n(t)+1\right)^{-(2\alpha_n^{(r)}-1)}\label{obj0} \\
\text{s.t.} \ \  &  \sum_{t \in \mathcal{T}^{(r)}} e_n(t)+\xi_{n}^{(r)} \leq E^{\text{cons}}_n,\quad \forall n \in \mathcal{N}^{(r)},  \label{energy1}\\
& s_{n,z}(t)=0,\quad p_{n,z}(t)=0,\label{long1} \\
&\forall n\in\mathcal N^{(r)},\ \forall z\in\mathcal Z,\ \forall t\in\mathcal T^{(r)},\
(t-1)\tau < \rho_n^{(r)},\notag\\
&\text{constraints (\ref{trans1})$-$(\ref{power1}), (\ref{fullupload}),}\notag
\end{align}
\end{subequations}
where $\boldsymbol{S}^{(r)} = [\boldsymbol{s}(1),...,\boldsymbol{s}(T)]$ denotes the RB allocation strategy, and $\boldsymbol{P}^{(r)} = [\boldsymbol{p}(1),...,\boldsymbol{p}(T)]$ denotes the power allocation strategy. Constraint (\ref{energy1}) ensures that the total energy consumption of each vehicle does not exceed the given energy budget. Constraint \eqref{long1} guarantees that uplink transmission starts only after local training is completed. Constraints \eqref{trans1}$-$\eqref{power1} specify the feasible transmission and power allocation regions.

\section{Progressive Gradient Transmission Scheme}
\begin{figure*}[t]  
    \centering
    \includegraphics[width=0.98\textwidth]{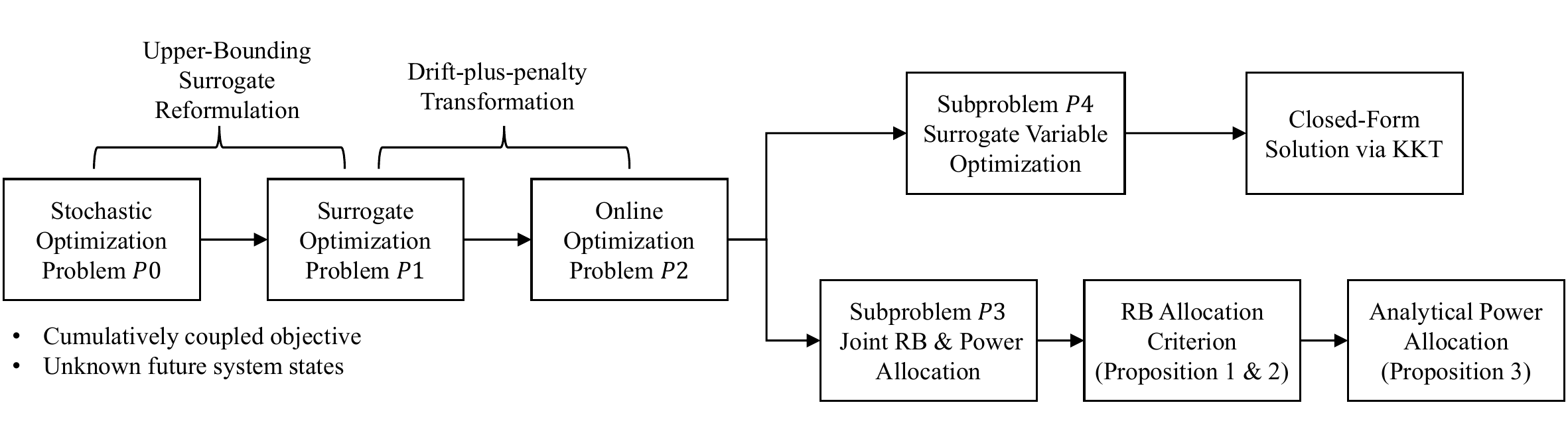}
    \caption{Optimization framework of the proposed FedPGT scheme.}
    \label{solution}
\end{figure*}

In this section, we present the proposed progressive gradient transmission scheme. We first employ a Lyapunov drift-plus-penalty approach to transform the original stochastic optimization problem into an online mixed-integer nonlinear programming (MINLP) problem. Then, we derive key structural properties that facilitate the design of a joint RB and power allocation algorithm. Based on these results, the proposed algorithm performs utility-driven RB assignment and analytical power allocation for efficient online implementation. The overall optimization framework is illustrated in Fig.~\ref{solution}.

\subsection{Transformation of the Stochastic Optimization Problem}
$P0$ is a stochastic optimization problem. The primary challenge in solving this problem lies in the uncertainty of future channel states and vehicle availability. High vehicle mobility results in rapidly varying communication conditions and unpredictable system evolution. Moreover, even with perfect future channel states and vehicle availability, solving $P0$ remains challenging due to the presence of integer and coupled decision variables.

Lyapunov optimization is an efficient approach for stochastic optimization with unknown future channel states and vehicle availability and enables online decision-making based solely on current observations. However, standard Lyapunov frameworks are designed for objectives that are additively separable over time, such as the minimization of long-term average costs or the maximization of time-average utilities. In contrast, the objective of $P0$ is \emph{cumulatively coupled}, depending on the sum of per-slot variables over the entire time horizon and preventing direct application of Lyapunov methods. To address this issue, we construct an upper-bounding surrogate reformulation that decouples the long-term dependence across time slots and converts the original cumulatively coupled objective into an additively separable surrogate objective, enabling Lyapunov-based per-slot optimization. The resulting surrogate optimization problem is given by
\begin{subequations}
\begin{align}
&\underset{\boldsymbol{S}^{(r)},\boldsymbol{P}^{(r)},\boldsymbol{\gamma}^{(r)}}{P1: \min}  \sum_{t\in\mathcal{T}^{(r)}}\sum_{n\in\mathcal{N}^{(r)}}\frac{(C_n^{(r)})^2\alpha_n^{(r)}\left(\gamma_n(t)+\frac{1}{T}\right)^{-(2\alpha_n^{(r)}-1)}}{2 \alpha_n^{(r)}-1}\label{obj1} \\
 &\text{s.t.} \ \ 0\leq\gamma_n(t)\leq\frac{I}{T},\quad \forall n \in \mathcal{N}^{(r)},\label{gammacon}\\ & \quad \ \ \sum_{t\in\mathcal{T}^{(r)}}\gamma_n(t) = \sum_{t\in\mathcal{T}^{(r)}}\kappa_n(t), \quad \forall n \in \mathcal{N}^{(r)}, \label{equ}\\
&  \quad \ \  \text{constraints \eqref{trans1}$-$\eqref{power1}, \eqref{fullupload}, \eqref{energy1}, \eqref{long1}}.\notag
\end{align}
\end{subequations}
In $P1$, the surrogate variable $\gamma_n(t)$ represents the slot-wise contribution to the cumulative transmitted gradient entries. The objective of $P1$ serves as an upper-bounding surrogate of the original cumulatively coupled objective in $P0$, while the equality constraint \eqref{equ} ensures long-term consistency between the surrogate variables and the actual transmitted gradient entries. 

To handle the long-term constraints in the surrogate optimization problem, we introduce two types of virtual queues. First, for each vehicle $n\in\mathcal N^{(r)}$, the long-term energy constraint \eqref{energy1} is enforced through the following virtual energy queue:
\begin{equation}
q_n(t+1)
=
\max\!\left[
q_n(t)+e_n(t)-\frac{E_n^{\mathrm{cons}}-\xi_n^{(r)}}{T},\,0
\right].
\end{equation}

The queue backlog $q_n(t)$ measures the accumulated energy deficit relative to the long-term energy budget. A larger queue backlog indicates that the vehicle has consumed excessive energy in previous slots, thereby imposing a stronger penalty on future transmission decisions.

Second, to enforce the equality constraint \eqref{equ}, we introduce a PGT queue defined as
\begin{equation}
\zeta_n(t+1)=\zeta_n(t)+\kappa_n(t)-\gamma_n(t),
\end{equation}
where the queue backlog $\zeta_n(t)$ measures the accumulated discrepancy between the actual transmitted gradient entries and the surrogate transmission variables. A larger queue backlog indicates that the number of transmitted entries exceeds the surrogate allocation, reducing the incentive to further transmit due to diminishing marginal convergence gains. Stabilizing this queue ensures long-term consistency between the surrogate reformulation and the original transmission process.

By applying the Lyapunov drift-plus-penalty framework to the surrogate optimization problem $P1$, we obtain the following per-slot optimization problem:
\begin{subequations}
\begin{align}
P2&:\underset{\boldsymbol s(t),\boldsymbol p(t),\boldsymbol\gamma(t)}{\min}
\sum_{n\in\mathcal N^{(r)}}\frac{V(C_n^{(r)})^2\alpha_n^{(r)}\left(\gamma_n(t)+\frac{1}{T}\right)^{-(2\alpha_n^{(r)}-1)}}{2 \alpha_n^{(r)}-1}\notag \\&
+\sum_{n\in\mathcal N^{(r)}} \left( q_n(t) e_n(t)
+\zeta_n(t) \kappa_n(t)  - \zeta_n(t)\gamma_n(t)\right)
\\
\text{s.t.} &\ \ \text{constraints \eqref{trans1}$-$\eqref{power1}, \eqref{fullupload}, \eqref{long1}, \eqref{gammacon}.}\notag
\end{align}
\end{subequations}
$P2$ is separable with respect to $\big(\boldsymbol s(t),\boldsymbol p(t)\big)$ and $\boldsymbol\gamma(t)$, since $\kappa_n(t)$ and $e_n(t)$ depend only on $\big(\boldsymbol s(t),\boldsymbol p(t)\big)$, while $\gamma_n(t)$ appears only in the terms $\frac{V(C_n^{(r)})^2\alpha_n^{(r)}\left(\gamma_n(t)+\frac{1}{T}\right)^{-(2\alpha_n^{(r)}-1)}}{2 \alpha_n^{(r)}-1}-\zeta_n(t)\gamma_n(t)$. Therefore, $P2$ can be decomposed into the following two subproblems. 
$P3$ determines the RB and power allocation decisions for the current slot:
\begin{align}
&P3: \underset{\boldsymbol s(t),\,\boldsymbol p(t)}{\min}\quad
\sum_{n\in\mathcal N^{(r)}} \left(q_n(t)\, e_n(t)
+  \zeta_n(t)\, \kappa_n(t)\right)
\\
&\text{s.t.}\quad
\text{constraints \eqref{trans1}$-$\eqref{power1}, \eqref{fullupload} \eqref{long1}.}\notag
\end{align}
The auxiliary variable $\boldsymbol \gamma(t)$ is optimized separately in $P4$:
\begin{align}
&P4: \underset{\boldsymbol \gamma(t)}{\min}
\sum_{n\in\mathcal N^{(r)}}\frac{V(C_n^{(r)})^2\alpha_n^{(r)}\left(\gamma_n(t)+\frac{1}{T}\right)^{-(2\alpha_n^{(r)}-1)}}{2 \alpha_n^{(r)}-1}\notag\\&\quad \quad \quad -\sum_{n\in\mathcal N^{(r)}}\zeta_n(t)\gamma_n(t)
\\
&\text{s.t.}\ \
\text{constraints \eqref{gammacon}}.\notag
\end{align}

Based on the above decomposition, we establish the following theorem to characterize the performance of the proposed Lyapunov-based online optimization framework. Superscript $^\dagger$ denotes the solution obtained by the online algorithm (i.e., solving $P3$ and $P4$ in every slot), while $^*$ denotes the optimal offline solution to $P0$.

\noindent \textbf{Theorem 2.} \emph{Suppose that all virtual queues are initialized to zero. Then, the performance gap between the solution obtained by the online algorithm and the optimal offline solution of $P0$ is bounded as}
\begin{equation}
\begin{aligned}
&\sum_{n\in\mathcal{N}^{(r)}}\frac{(C_n^{(r)})^2\alpha_n^{(r)}}{2 \alpha_n^{(r)}-1}\bigg(\sum_{t\in\mathcal{T}^{(r)}}\kappa_n^\dagger(t)+1\bigg)^{-(2\alpha_n^{(r)}-1)} \\& \leq \sum_{n\in \mathcal{N}^{(r)}} \frac{(C_n^{(r)})^2\alpha_n^{(r)}(k_n^{(r)*}+1)^{-(2\alpha_n^{(r)}-1)}}{2 \alpha_n^{(r)}-1}\\&+\sum_{n\in \mathcal{N}^{(r)}}\frac{T^{2-2\alpha_n^{(r)}} \Phi}{V}+(C_n^{(r)})^2\alpha_n^{(r)} T \sqrt{2\Phi}.
\end{aligned}
\end{equation}
\emph{The energy consumption of vehicle $n$ is bounded by}
\begin{equation}
\begin{aligned}
&\xi_n^{(r)}+\sum_{t\in\mathcal{T}^{(r)}}  e_{n}^\dagger(t)   \leq E^{\text{cons}}_n+ T\sqrt{2 \Phi},
\end{aligned}
\end{equation}
\emph{where $\epsilon_{n}(t) \triangleq  e_{n}(t)- \frac{E^{\text{cons}}_n-\xi_n^{(r)}}{T}$, $\phi_n \triangleq \max_t\{|\epsilon_{n}(t)|\}$, $\varphi_n\triangleq\max_t\{|\kappa_n(t)-\gamma_n(t)|\}$ and $\Phi \triangleq\max_n\{(\phi_n)^2+(\varphi_n)^2\}$.}

\noindent \emph{Proof:} See Appendix \ref{Theorem2}. \hfill$\square$

Theorem~2 characterizes the performance of the Lyapunov-based online framework under exact per-slot minimization of $P3$ and $P4$. Specifically, if the decomposed per-slot problems are solved optimally in each slot, the resulting online policy achieves a bounded gap with respect to the offline optimum of $P0$, while guaranteeing bounded long-term energy consumption for each vehicle.

Based on the above decomposition, we next solve the two subproblems $P3$ and $P4$, respectively.

$P4$ is a convex optimization problem, since its objective function is convex with respect to $\boldsymbol\gamma(t)$ and the feasible set defined by \eqref{gammacon} is convex. Therefore, the optimal solution can be directly obtained from the KKT conditions as
\begin{equation}
\gamma_n^{\dagger}(t)
=
\begin{cases}
\frac{I}{T}, & \zeta_n(t)\ge 0,\\
\left[
\left(\frac{ V (C_n^{(r)})^2\alpha_n^{(r)}}{-\zeta_n(t)}\right)^{\frac{1}{2\alpha_n^{(r)}}}-\frac{1}{T}
\right]^{\frac{I}{T}}_0, & \zeta_n(t)<0,
\end{cases}
\end{equation}
where $[a]^{\frac{I}{T}}_0$ is defined as $\min(\max(a,0), \frac{I}{T})$.

In contrast, $P3$ is a MINLP problem due to the binary RB allocation variables and the nonlinear coupling introduced by the power control constraints.  As a result, solving $P3$ is more challenging and will be addressed in the following subsections.

\subsection{RB Allocation Criterion}

First, we relax the binary RB allocation variables $\boldsymbol{s}(t)$ to continuous variables in $[0,1]$ and consider the following relaxed version of $P3$:
\begin{subequations}
\begin{align}
P3:  &\underset{\boldsymbol{s}(t),\boldsymbol{p}(t)}{\min} \sum_{n\in\mathcal{N}(t)}  \sum_{z\in\mathcal{Z}}  \frac{\zeta_n(t)\tau  \beta s_{n,z}(t) \log_2 (1+ \Gamma_{n,z}(t))}{b+\lceil \log_2 I \rceil}\notag\\&\  \ \ \ \ \ +\sum_{n\in\mathcal{N}(t)}  \sum_{z\in\mathcal{Z}} q_n(t) \tau s_{n,z}(t) p_{n,z}(t) \label{objp2}\\
\text{s.t.} \ 
&  0\leq s_{n,z}(t) \leq1, \quad \forall n \in \mathcal{N}(t), \ \forall z \in \mathcal{Z}, \label{relaxed}\\
& \text{constraints (\ref{trans1}), (\ref{power1})},\notag
\end{align}
\end{subequations}
where $\mathcal{N}(t)=\{n\in\mathcal{N}^{(r)}:(t-1)\tau\geq\rho_n^{(r)},\sum_{t'=1}^{t-1}\kappa_n(t')< I\}$ denotes the set of vehicles that have completed local training and are eligible for transmission at slot t. Owing to constraint \eqref{long1}, vehicles outside $\mathcal{N}(t)$ are excluded from RB allocation. The relaxed problem $P3$ remains non-convex since the objective function is neither convex nor concave with respect to $\boldsymbol{s}(t)$ and $\boldsymbol{p}(t)$. Thus, obtaining a globally optimum solution is challenging. Nevertheless, we derive the following proposition characterizing the necessary optimality conditions based on the KKT conditions. 

\noindent \textbf{Proposition 1.} \emph{A necessary condition for $s_{n,z}(t)$ being positive at the optimal solution is} 
\begin{equation}
n = \arg\min_{n\in\mathcal{N}(t)} \Upsilon_{n,z}(t)
\end{equation}
\emph{where the cost function $\Upsilon_{n,z}(t)$ is defined as}
\begin{equation}
\Upsilon_{n,z}(t) = \frac{\zeta_n(t)\tau \beta \log_2 (1+ \Gamma_{n,z}(t))}{b+\lceil \log_2 I \rceil}  + \tau q_{n}(t) p_{n,z}(t).\label{U}
\end{equation}
\noindent \emph{Proof:} See Appendix \ref{Proposition1}. \hfill$\square$

This proposition characterizes a necessary optimality condition for RB allocation. Although it does not directly yield the globally optimal solution, it provides the basis for the proposed RB allocation criterion.

Although the binary RB allocation variables are relaxed to continuous variables in $[0,1]$, the following proposition shows that, for a given power allocation, the relaxed RB allocation subproblem admits an integral optimal solution. Therefore, the exclusive RB allocation property of OFDMA is preserved without loss of optimality.

\noindent \textbf{Proposition 2.}
\emph{For a given power allocation, the relaxed RB allocation subproblem admits an integral optimal solution. Specifically, there exists an optimal solution in which each RB is exclusively assigned to one vehicle, i.e.,}
\[
s_{n^*,z}(t)=1,
\quad
s_{n,z}(t)=0,
\quad
\forall n\neq n^*.
\]

\noindent \emph{Proof:} Consider a fixed RB $z$. Let $\mathcal{N}_{z}(t)$ denote the set of vehicles that achieve the smallest $\Upsilon_{n,z}(t)$ on RB $z$, i.e.,
$$\mathcal{N}_{z}(t) = \left\{ n \,|\, \arg\min_{n\in\mathcal{N}(t)} \Upsilon_{n,z}(t) \right\}.$$
From the KKT conditions, we know that for the optimal solution, $s_{n,z}(t) > 0$ implies user $n \in \mathcal{N}_{z}(t)$, and $s_{n,z}(t) = 0$ for all $n \notin \mathcal{N}_{z}(t)$. When RB $z$ is allocated to a certain vehicle, constraint \eqref{trans1} becomes
$\sum_{n \in \mathcal{N}(t)} s_{n,z}(t) = \sum_{n \in \mathcal{N}_{z}(t)} s_{n,z}(t) = 1,$
Moreover, from the definition of $\mathcal{N}_{z}(t)$, all vehicles $n \in \mathcal{N}_{z}(t)$ have the same value of $\Upsilon_{n,z}(t)$, which is the smallest possible among all vehicles. As a result, the contribution of the objective (\ref{objp2}) from RB $z$ is
$$
\sum_{n \in \mathcal{N}_{z}(t)} s_{n,z}(t) \Upsilon_{n,z}(t) = \Upsilon_{n^*,z}(t) \sum_{n \in \mathcal{N}_{z}(t)}s_{n,z}(t)= \Upsilon_{n^*,z}(t),
$$
where $n^* \in \mathcal{N}_{z}(t)$.

\begin{algorithm}[!t]	
    \caption{Power allocation algorithm} 
    \label{poweralgo}
	\begin{algorithmic}
        \STATE{Sort $z \in \mathcal{Z}_n(t)$ in descending order of $\big\Vert \boldsymbol{h}_{n,z}(t) \big\Vert^2$.}
	\FOR{$j = 1$ \textbf{to} $|\mathcal{Z}_n(t)|$}
        \STATE{Select the first $j$ RBs as $\mathcal{Z}_n^*(t)$;}
        \STATE{Compute power allocation $p'_{n,z}(t)$ for all $z \in \mathcal{Z}_n^*(t)$ according to Proposition 3;}
        \IF{any $p'_{n,z}(t) \le 0$}
        \STATE{\textbf{break};}
        \COMMENT{Stop: invalid RB found, do not update}
        \ELSE
        \STATE{Update the current power allocation $p_{n,z}(t) = p'_{n,z}(t)$;}
        \ENDIF
        \ENDFOR
	\end{algorithmic}
\end{algorithm}

Therefore, regardless of how the weights
$s_{n,z}(t)$
are distributed among vehicles in
$\mathcal N_z(t)$, the objective contribution remains unchanged as long as their sum equals 1. 

Furthermore, $\Upsilon_{n^*,z}(t) \leq 0$ holds because the transmit power $p_{n^*,z}(t)$ associated with the RB can be set to zero without violating any constraints. Therefore, assigning the RB to a vehicle is no worse than leaving it unassigned.

Hence, for the relaxed RB allocation subproblem with fixed power allocation, there exists an integral solution assigning RB $z$ exclusively to one vehicle. This completes the proof. \hfill$\square$

\subsection{Power Allocation}
Then, the power allocation problem is considered. Given the RB allocation decision, the following proposition characterizes the optimal transmit power allocation strategy.

\noindent \textbf{Proposition 3.} \emph{Given the RB allocation decision, the optimal power allocation for vehicle $n$ is characterized as follows.} 

\emph{If $\zeta_n(t)\ge 0$, the optimal transmit power is}
\begin{equation}
p_{n,z}(t)=0,\quad \forall z\in \mathcal Z_n(t).
\end{equation}

\emph{If $\zeta_n(t)<0$ and $q_n(t)>0$, the optimal transmit power over the positive-power RB set $\mathcal Z_n^*(t)$ is}
\begin{align}p_{n,z}(t) &=  \min \left(\frac{p^{\text{max}}_n+\sum_{z'\in\mathcal{Z}_{n}^*(t)}\frac{\beta N_0}{\big\Vert \boldsymbol{h}_{n,z'}(t) \big\Vert^2}}{\vert \mathcal{Z}_{n}^*(t)\vert}, \frac{-\zeta_n(t) \beta}{q_{n}(t)\ln2}  \right)\notag\\
&- \frac{\beta N_0}{\big\Vert \boldsymbol{h}_{n,z}(t)\big\Vert^2}, \quad \forall z \in \mathcal{Z}_{n}^*(t).\label{powerallocation}\end{align}

\emph{If $\zeta_n(t)<0$ and $q_n(t)=0$, the optimal transmit power is}
\begin{equation}
\begin{aligned}
p_{n,z}(t)
=&
\frac{
p_n^{\max}
+
\sum_{z'\in\mathcal Z_n^*(t)}
\frac{\beta N_0}{\|\boldsymbol h_{n,z'}(t)\|^2}
}
{|\mathcal Z_n^*(t)|}
-
\frac{\beta N_0}{\|\boldsymbol h_{n,z}(t)\|^2},
\\ &\forall z\in\mathcal Z_n^*(t).
\end{aligned}
\end{equation}

\noindent \emph{Proof:} See Appendix \ref{Proposition3}. \hfill$\square$

Proposition 3 gives a closed-form expression for power allocation. However, this requires prior knowledge of the subset of RBs to which vehicle $n$ assigns positive power, i.e., $\mathcal{Z}_{n}^*(t)$. To address this, we propose Algorithm \ref{poweralgo}, which incrementally searches for the subset $\mathcal{Z}_{n}^*(t)$. Specifically, it first sorts all candidate RBs in descending order of channel gain. Then, it sequentially adds RBs to the candidate set and computes the corresponding power allocation using (\ref{powerallocation}). The process continues until a non-positive power value is encountered, at which point the search terminates. This ensures that only RBs receiving positive power are included in $\mathcal{Z}_{n}^*(t)$.

\noindent \textbf{Remark 2.} Propositions~1 and~3 characterize the structural properties of the optimal RB and power allocation strategies. The PGT queue $\zeta_n(t)$ regulates the transmission urgency according to the accumulated transmitted gradient entries. When $\zeta_n(t)\geq0$, the vehicle has already transmitted a relatively large number of gradient entries, and the diminishing marginal convergence gain reduces the incentive for further transmissions. In this case, the transmit power becomes zero, and the corresponding RBs are released for other vehicles with more urgent transmission demands. When $\zeta_n(t)<0$, the system allocates more communication resources to increase the transmitted gradient entries. The energy queue $q_n(t)$ controls the long-term energy consumption, where a larger queue backlog imposes a higher penalty on future power allocation and encourages more energy-efficient transmissions.

\begin{algorithm}[!t]	
    \caption{Joint RB and power allocation algorithm}
    \label{algo:global-wf}
	\begin{algorithmic}[1]
        \STATE{Initialize all $s_{n,z}(t) = 0$, $p_{n,z}(t) = 0$, set of unallocated RBs $\mathcal{Z}'(t) = \mathcal{Z}$, and the set of RBs already allocated to vehicle $n$ $\mathcal{Z}_n^\dagger(t)=\emptyset$;}
        \WHILE{$\mathcal{Z}' \neq \emptyset$}
            \FOR{$n \in \mathcal{N}(t)$}
                \STATE Let $\mathcal{Z}_n(t) = \mathcal{Z}_n^\dagger(t) \cup \mathcal{Z}'(t)$;
                \STATE Obtain the power allocation $p_{n,z}(t)$ over $\mathcal{Z}_n(t)$ based on Algorithm \ref{poweralgo};
                \STATE Compute $\Upsilon_{n,z}(t)$ according to (\ref{U});
            \ENDFOR
            \STATE Find $(n, z) = \arg\min_{n \in \mathcal{N}(t), z \in \mathcal{Z}'} \Upsilon_{n,z}(t)$;
            \STATE Allocate RB $z$ to user $n$: set $s_{n, z}(t) = 1$;
            \STATE Update $\mathcal{Z}' \leftarrow \mathcal{Z}' \setminus \{z\}$ and $\mathcal{Z}_n^\dagger(t) \leftarrow \mathcal{Z}_n^\dagger(t) \cup \{z\}$;
        \ENDWHILE
        \FOR{$n \in \mathcal{N}(t)$}
        \STATE Perform Algorithm \ref{poweralgo} over the assigned RBs $\{z | s_{n,z}(t) = 1\}$ to finalize $p_{n,z}(t)$;
        \ENDFOR
	\end{algorithmic}
\end{algorithm}

\subsection{Joint RB and Power Allocation Algorithm}
Although Propositions~1--3 provide analytical characterizations for the RB and power allocation strategies, RB allocation and power allocation remain inherently coupled, since the RB utility depends on the power allocation, while the power allocation itself depends on the allocated RB set. We develop the iterative joint RB and power allocation algorithm shown in Algorithm~\ref{algo:global-wf} for solving $P3$.

The proposed algorithm iteratively allocates RBs according to their marginal utility contributions under adaptive power allocation. Initially, all RBs are unallocated. In each iteration, for every vehicle $n$, the RSU temporarily augments its currently allocated RB set $\mathcal Z_n^\dagger(t)$ with the remaining unallocated RBs $\mathcal Z'(t)$. Based on this temporary RB set, the corresponding power allocation is recomputed using Algorithm~\ref{poweralgo}. The resulting $\Upsilon_{n,z}(t)$ value is then evaluated for all candidate RB-vehicle pairs.

\begin{algorithm}[!t]
\caption{The Overall Procedure of FedPGT}
\label{AlgoDSRA}
\begin{algorithmic}
\STATE{\textbf{Initialization:}
Set $\boldsymbol q(1)=\boldsymbol 0$ and
$\boldsymbol\zeta(1)=\boldsymbol 0$;}
\FOR{$t\in\mathcal T^{(r)}$}
    \STATE{Vehicles completing local training estimate
    $C_n^{(r)}$ and $\alpha_n^{(r)}$
    via power-law fitting;}
    \STATE{Construct the active transmission set
    $\mathcal N(t)$;}
    \STATE{Observe current CSI
    $\boldsymbol h(t)$
    and queue states
    $\boldsymbol q(t)$, $\boldsymbol\zeta(t)$;}
    \STATE{Solve $P4$ to obtain
    $\boldsymbol\gamma^*(t)$;}
    \STATE{Solve $P3$ to obtain $\boldsymbol{s}^*(t)$ and $\boldsymbol{p}^*(t)$ using Algorithm~\ref{algo:global-wf};}
    \STATE{Vehicles in $\mathcal N(t)$ transmit gradient entries;}
    \STATE{Update virtual queues
    $\boldsymbol q(t+1)$ and
    $\boldsymbol\zeta(t+1)$;}
\ENDFOR
\end{algorithmic}
\end{algorithm}

The RB-vehicle pair $(n,z)$ yielding the smallest $\Upsilon_{n,z}(t)$ is selected, and RB $z$ is assigned to vehicle $n$. The allocated RB is then removed from the unallocated RB set, and the procedure repeats until all RBs are assigned. Finally, after the RB allocation process is completed, each vehicle recomputes the power allocation over its finalized RB set using Algorithm~\ref{poweralgo}.

\subsection{The Overall Procedure of FedPGT}

The workflow of the proposed PGT algorithm is outlined in Algorithm \ref{AlgoDSRA}. At the start of each round, the RSU broadcasts the global model and vehicles conduct local model updates. The compressibility parameters $C_n^{(r)}$ and $\alpha_n^{(r)}$ are then estimated locally by fitting the sorted gradient magnitudes to the power-law model in the log-log domain. In each slot, the RSU observes the current channel states and vehicle availability and solves the decomposed per-slot problems $P3$ and $P4$. Vehicles progressively transmit gradient entries according to the obtained RB and power allocation decisions, and the virtual queues are updated based on the realized transmissions. This procedure continues over all slots within the round. By solving the per-slot problems online, PGT adapts to time-varying channels without requiring future system information and achieves a bounded performance gap.

\begin{figure*}[!t]
  \centering
  \begin{minipage}[t]{0.32\textwidth}
    \vspace{0pt} 
    \centering
    \includegraphics[width=\textwidth]{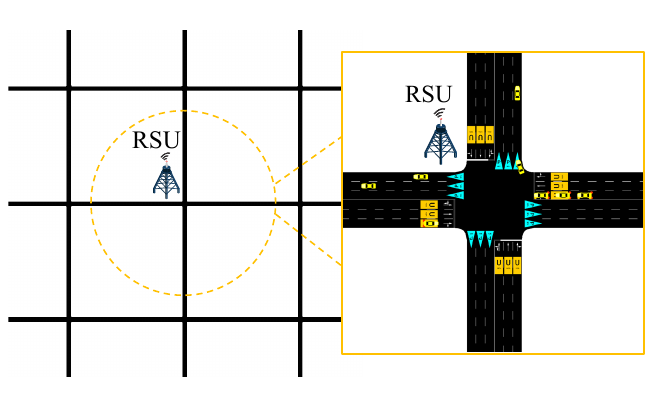}
    \caption{The road network generated by SUMO.}
    \label{SUMO}
  \end{minipage}
  \hfill
  \begin{minipage}[t]{0.32\textwidth}
    \vspace{0pt}
    \centering
    \includegraphics[width=\textwidth]{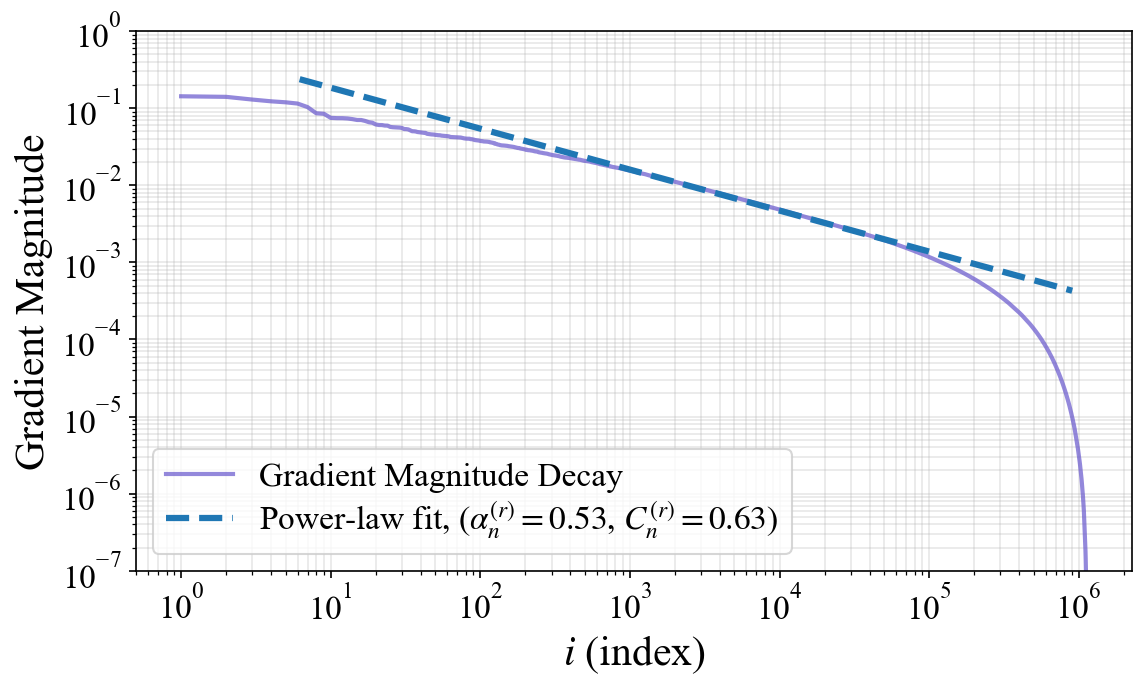}
    \caption{Sorted Gradient Magnitudes in the log-log Domain.}
    \label{gradient}
  \end{minipage}
  \hfill
  \begin{minipage}[t]{0.32\textwidth}
    \vspace{0pt}
    \centering
    \includegraphics[width=\textwidth]{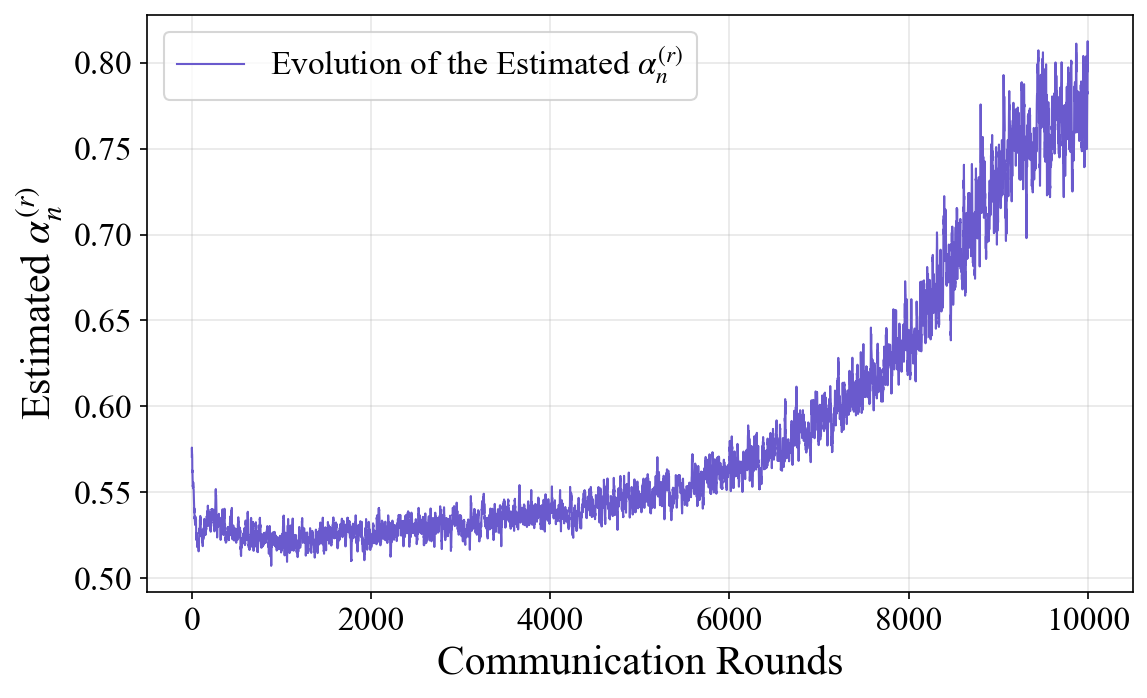}
    \caption{Evolution of Estimated Compressibility Exponent $\alpha_n^{(r)}$.}
    \label{alpha}
  \end{minipage}
    \hfill
\end{figure*}

The per-slot complexity is dominated by solving $P3$ and $P4$. $P4$ admits a closed-form solution, and its complexity scales linearly with the number of vehicles, i.e., $\mathcal O(N^{(r)})$. For $P3$, the RB assignment iterates over all unallocated RBs. In each iteration, Algorithm~\ref{poweralgo} is executed for each vehicle to recompute the corresponding power allocation, yielding complexity $\mathcal{O}(N^{(r)}Z)$ per iteration. Since each RB is assigned once, the overall complexity of the RB allocation is $\mathcal{O}(N^{(r)}Z^2)$. Note that Algorithm~\ref{poweralgo} involves sorting RBs according to their channel gains. However, this sorting operation is only performed once for each vehicle at the beginning of each slot, and the resulting RB ordering is reused throughout subsequent executions. Therefore, the sorting step only introduces preprocessing complexity $\mathcal{O}(N^{(r)}Z\log_2 Z)$, which is dominated by $\mathcal{O}(N^{(r)}Z^2)$.

Therefore, the per-slot complexity is $\mathcal{O}(N^{(r)}Z^2)$. Over $T$ slots, the total complexity becomes $\mathcal{O}(TN^{(r)}Z^2)$.

\begin{table}[t!]
\caption{Simulation Parameters.}
\begin{tabular}{l|l}
\hline
\textbf{Simulation Parameters}                   & \textbf{Values}                            \\ \hline
System Bandwidth             & 20 MHz       
\\ \hline
Maximum Transmission Power   & 0.2 W                              \\ \hline
Carrier Frequency            & 5.9 GHz                            \\ \hline
Vehicle Blockage Loss        & $\max\{0, \mathcal{N}(5, 4)\}$ dB \\ \hline
Shadowing Fading Std. Dev.   & 3 dB (LOS, NLOSv), 4 dB (NLOS)      \\ \hline
Noise Power Spectrum Density & -174 dBm/Hz                        \\ \hline
Energy Consumption Coefficient      & $10^{-28}$ \cite{energyco1, energyco2}\\ \hline
Energy Constraints        & Randomly selected from $0.05$ J to $0.1$ J\\ \hline
Length of Time Slot & $10$ ms\\ \hline
Average Number of Vehicles & 15 \\ \hline
Number of RBs & 50 \\ \hline
\end{tabular}
\label{table}
\end{table}

\section{Experimental Setup}
In this section, we evaluate the proposed FedPGT through simulations under vehicular network settings. We first introduce the simulation setup. Then, the performance of FedPGT under different vehicular network conditions is evaluated on both image classification and trajectory prediction tasks using the CIFAR-10 and Argoverse datasets.

\subsection{Simulation Setups}

An urban grid road network is generated using SUMO \cite{SUMO}, where an RSU is deployed at the center. Vehicle routing follows a Manhattan mobility pattern. At each intersection, vehicles move straight with probability $0.5$ and turns left or right with probability $0.25$ each.

Vehicle dynamics follow the Intelligent Driver Model (IDM). The acceleration of vehicle $n$ is given by
\begin{equation}
a_n = a^{\text{max}}\!\left[1 - \left(\frac{v_n}{v^{\text{max}}}\right)^4 - \left(\frac{d'_n}{d^*_n}\right)^2 \right],\notag
\end{equation}
where $v_n$ is the instantaneous speed, $v^{\text{max}}$ is the maximum allowable speed, and $d'_n$ is the headway distance.

The desired headway distance is given by
\begin{equation}
d^*_n = d^{\text{safe}} + v_n t^{\text{dst}} + \frac{v_n \Delta v_n}{2\sqrt{a^{\text{max}} a^{\text{brake}}}},\notag
\end{equation}
where $d^{\text{safe}}$ is the safety distance, $t^{\text{dst}}$ is the desired time headway, $\Delta v_n$ is the relative speed, and $a^{\text{brake}}$ denotes the comfortable braking deceleration.

The wireless channel model follows the 3GPP V2X specifications in TR 37.885 \cite{V2Vstd2}. Both line-of-sight (LOS) and non-line-of-sight (NLOS) propagation conditions are considered.

The urban LOS path loss is modeled as
$PL_{\text{LOS}} = 38.77 + 16.7\log_{10}(d) + 18.2\log_{10}(f_c)$,
and the NLOS path loss is
$PL_{\text{NLOS}} = 36.85 + 30\log_{10}(d) + 18.9\log_{10}(f_c)$,
where $d$ is the link distance and $f_c$ is the carrier frequency. The shadowing coefficient follows a log-normal distribution, while the small-scale fading vector follows Rayleigh fading.

The key simulation parameters are summarized in Table~\ref{table}. Unless otherwise specified, the default parameter settings are adopted when evaluating the impact of individual variables.

\subsection{Model and Datasets}

We evaluate the proposed FedPGT on both image classification and trajectory prediction tasks.

\subsubsection{CIFAR-10} We first evaluate FedPGT on the CIFAR-10 dataset \cite{cifar10}, which contains 50,000 training images and 10,000 test images from 10 classes. A non-i.i.d. data distribution is considered, where samples are partitioned according to labels and each vehicle holds data from two distinct classes.

A convolutional neural network is trained for image classification. The network consists of six convolutional layers, followed by ReLU activations, with alternating max-pooling and normalization operations. The final convolutional layer is connected to a fully connected layer and a softmax classifier. The batch size is randomly selected from \{16, 32, 48\}, and the learning rate is set to 0.1.

\subsubsection{Argoverse} We further evaluate FedPGT on the Argoverse trajectory prediction dataset \cite{Argoverse}, which contains over 300,000 sequences collected in urban driving scenarios. Each sequence is sampled at 10 Hz, and the task is to predict future positions over a 3-second horizon. The dataset is divided into training, validation, and test sets with 205,942, 39,472, and 78,143 sequences, respectively. The data are uniformly partitioned into 40 subsets.

For trajectory prediction, we adopt the LaneGCN model \cite{lane}, which integrates trajectory encoding and map information. It consists of an ActorNet for extracting trajectory features, a MapNet for modeling lane graph structures, and a FusionNet for combining these features to generate predictions. The batch size is randomly selected from \{16, 32, 48\}, and the learning rate is set to 0.1. We use the average displacement error (ADE) as the evaluation metric.

\begin{figure*}[!t]
  \centering
  \begin{minipage}[t]{0.32\textwidth}
    \vspace{0pt} 
    \centering
    \includegraphics[width=\textwidth]{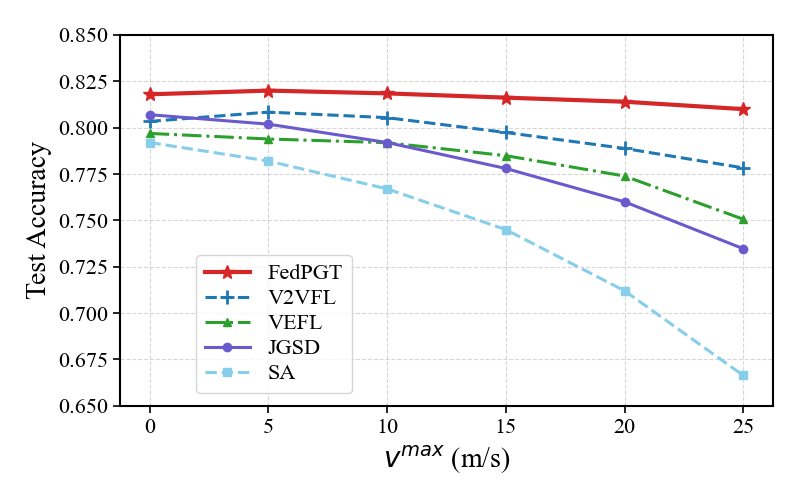}
    \caption{Final test accuracy of different methods under different vehicle speeds.}
    \label{speed}
  \end{minipage}
  \hfill
  \begin{minipage}[t]{0.32\textwidth}
    \vspace{0pt}
    \centering
    \includegraphics[width=\textwidth]{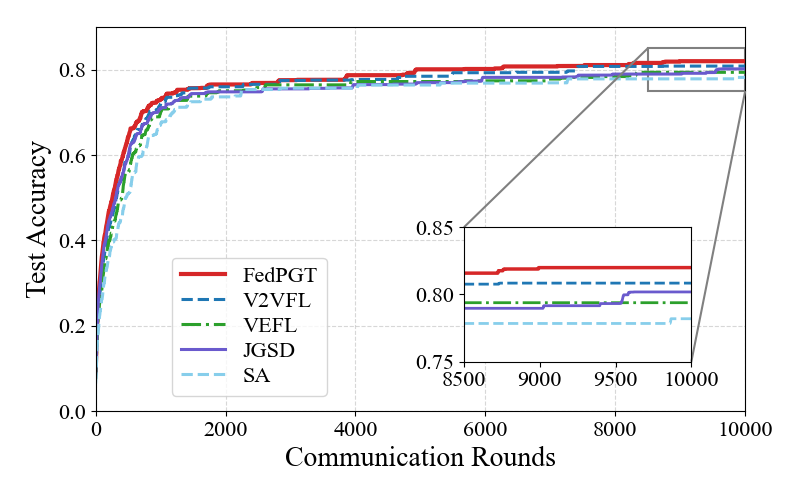}
    \caption{Convergence performance of different methods when $v^{\max}=5$ m/s.}
    \label{speedlow}
  \end{minipage}
  \hfill
  \begin{minipage}[t]{0.32\textwidth}
    \vspace{0pt}
    \centering
    \includegraphics[width=\textwidth]{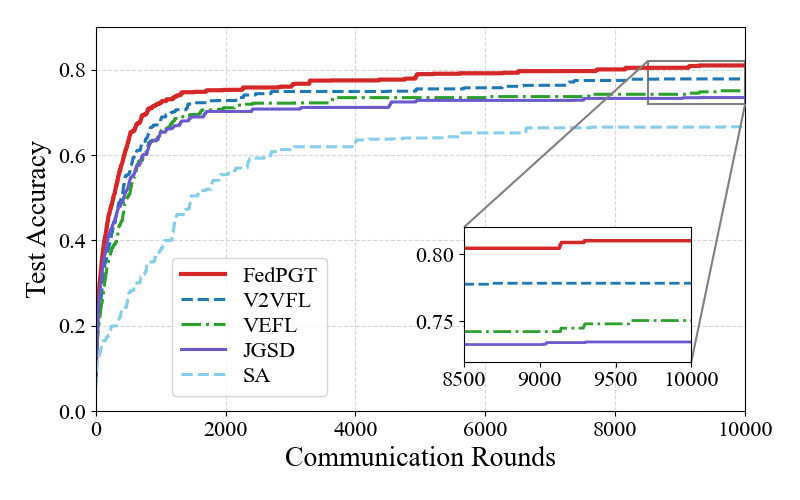}
    \caption{Convergence performance of different methods when $v^{\max}=25$ m/s.}
    \label{speedhigh}
  \end{minipage}
    \hfill
  \begin{minipage}[t]{0.32\textwidth}
    \vspace{0pt}
    \centering
    \includegraphics[width=\textwidth]{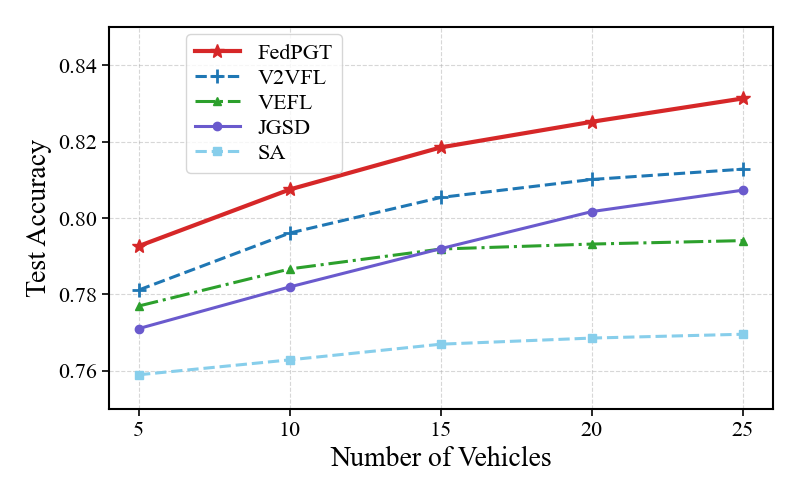}
    \caption{Final test accuracy of different methods under different average number of vehicles.}
    \label{number}
  \end{minipage}
  \hfill
    \begin{minipage}[t]{0.32\textwidth}
    \vspace{0pt} 
    \centering
    \includegraphics[width=\textwidth]{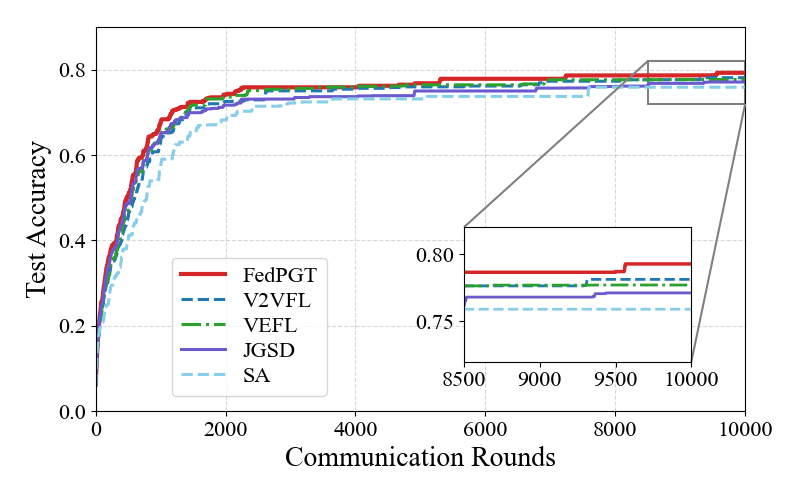}
    \caption{Convergence performance when the average number of vehicles is $5$.}
    \label{number5}
  \end{minipage}
  \hfill
  \begin{minipage}[t]{0.32\textwidth}
    \vspace{0pt}
    \centering
    \includegraphics[width=\textwidth]{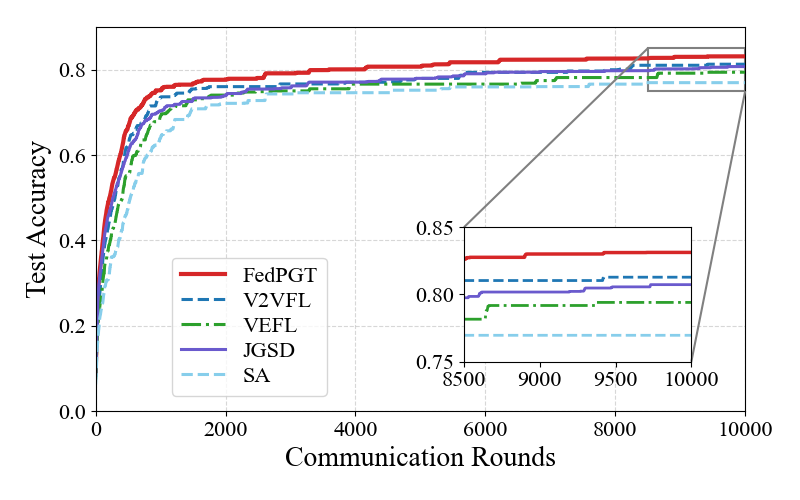}
    \caption{Convergence performance when the average number of vehicles is $25$.}
    \label{number25}
  \end{minipage}
\end{figure*}

\subsection{Baseline Schemes}

To evaluate the performance of the proposed FedPGT, we compare it with the following benchmark schemes.

\subsubsection{V2V-Enhanced FL (V2VFL) \cite{Yan2025DynamicSchedulingV2VFL}}

This scheme adopts dynamic device scheduling with V2V-assisted communications. Transmission and scheduling decisions are updated at each time slot according to channel variations and vehicle mobility. Unlike FedPGT, only fully uploaded updates can contribute to the model aggregation.

\subsubsection{Vehicular Edge FL (VEFL) \cite{VFL8}}

VEFL considers vehicle mobility and fast channel variations. Resource allocation decisions are dynamically updated according to instantaneous channel states.

\subsubsection{Joint Gradient Sparsification and Device Scheduling (JGSD) \cite{Lin2023JGSD}}

JGSD jointly considers device scheduling and gradient sparsification, where the sparsification degree is fixed at the beginning of each communication round.

\subsubsection{Static Scheduling Algorithm (SA) \cite{DS7}}

This scheme performs device scheduling and resource allocation based on initial channel states and vehicle locations, without adapting to mobility or applying sparsification.

\subsection{Estimation of Gradient Compressibility Parameters}

In this subsection, we illustrate how the compressibility parameters are estimated from the local gradients and investigate their evolution during federated training. According to Definition~1 and Assumption~1, 
$|g_{n,i}^{(r)}|
\le C_n^{(r)} i^{-\alpha_n^{(r)}}$,
which implies that the sorted gradient magnitudes are upper-bounded by a linear function in the log-log domain:
$\log_{10} |g_{n,i}^{(r)}| \le \log_{10} C_n^{(r)} - \alpha_n^{(r)} \log_{10} i$. This motivates the estimation of the compressibility parameters through linear regression in the log-log domain.

Fig.~\ref{gradient} illustrates the sorted gradient magnitudes. The dominant gradient entries exhibit an approximately linear relationship in the log-log domain, indicating that a power-law model provides a good approximation of the gradient decay behavior. In contrast, the tail entries decrease more rapidly than the fitted power-law curve, primarily due to the large number of near-zero gradient components. Since Assumption~1 only requires the sorted gradient magnitudes to be upper-bounded by a power-law function, such a faster decay remains consistent with the assumption. To estimate the compressibility parameters, we identify the dominant region that exhibits the strongest linear relationship in the log-log domain. A sliding-window least-squares regression is performed over the sorted gradient magnitudes, and the interval achieving the highest coefficient of determination value is selected for parameter estimation. The resulting fitting parameters are then used as the scaling coefficient $C_n^{(r)}$ and compressibility exponent $\alpha_n^{(r)}$ in the proposed framework.

Fig.~\ref{alpha} shows the evolution of the estimated compressibility exponent $\alpha_n^{(r)}$ during training. As training progresses, the estimated exponent gradually increases, indicating that the gradients become increasingly compressible near convergence. This is because the gradient magnitude becomes progressively concentrated on a few dominant entries as more model parameters approach stationary points. Consequently, fewer dominant gradient entries are required to accurately approximate the full gradient, further demonstrating the effectiveness of progressive gradient transmission.

\subsection{Performance under Different Vehicle Speeds}

In this subsection, we evaluate performance under varying maximum vehicle speeds $v^{\text{max}}$. Fig.~\ref{speed} shows the final accuracy, while Figs.~\ref{speedlow} and \ref{speedhigh} present the convergence performance at $v^{\text{max}}=5$ m/s and $v^{\text{max}}=25$ m/s, respectively.

At low vehicle speeds, all schemes achieve comparable performance, as vehicles remain within the RSU coverage long enough to complete transmissions. Compared with the static case, FedPGT and V2VFL even achieve slight gains due to channel diversity introduced by mobility. When vehicles move at high speeds, performance degrades across all schemes. Higher mobility shortens the sojourn time and increases the probability of interrupted transmissions, reducing the number of successfully aggregated updates.

Among all schemes, FedPGT exhibits the strongest robustness against mobility, as it jointly incorporates dynamic resource allocation and progressive gradient transmission. Dynamic resource allocation adapts to time-varying channel conditions, while progressive gradient transmission allows partially transmitted updates to contribute to model aggregation under interrupted transmissions.

V2VFL and VEFL perform well at low speeds but degrade at high speeds. These schemes require full model uploads, and thus transmission interruptions caused by mobility lead to invalid updates. JGSD also degrades with increasing speed, but remains more robust than SA because partially transmitted model updates can still contribute to training. In contrast, SA suffers the most severe degradation, as it neither adapts to channel variations nor accounts for vehicle mobility.

\begin{figure*}[!t]
  \centering
  \begin{minipage}[t]{0.32\textwidth}
    \vspace{0pt} 
    \centering
    \includegraphics[width=\textwidth]{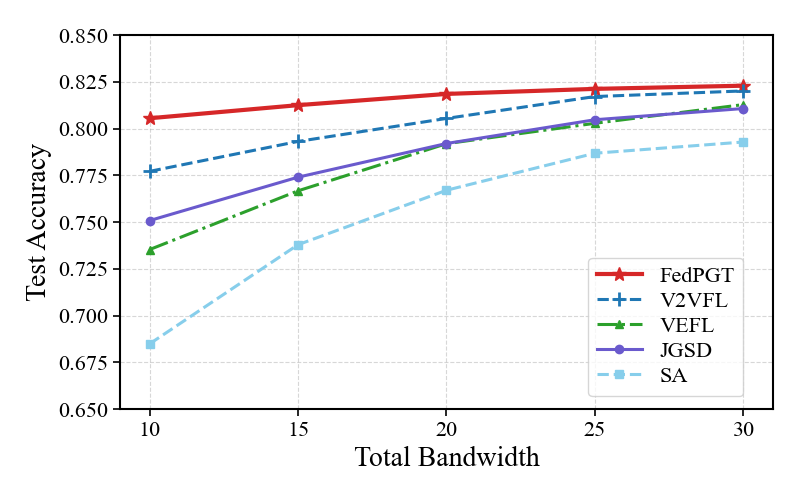}
    \caption{Final test accuracy of different methods under different total bandwidth.}
    \label{bandwidth}
  \end{minipage}
  \hfill
  \begin{minipage}[t]{0.32\textwidth}
    \vspace{0pt}
    \centering
    \includegraphics[width=\textwidth]{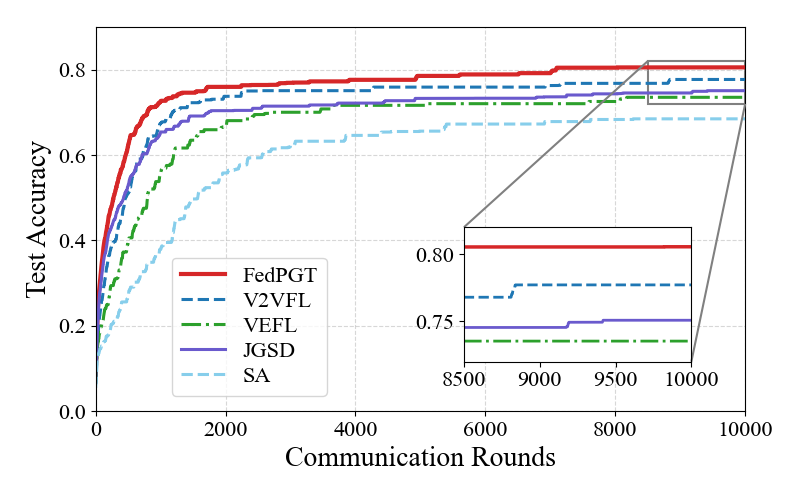}
    \caption{Convergence performance when the total bandwidth is $10$ MHz.}
    \label{bandwidth10}
  \end{minipage}
  \hfill
  \begin{minipage}[t]{0.32\textwidth}
    \vspace{0pt} 
    \centering
    \includegraphics[width=\textwidth]{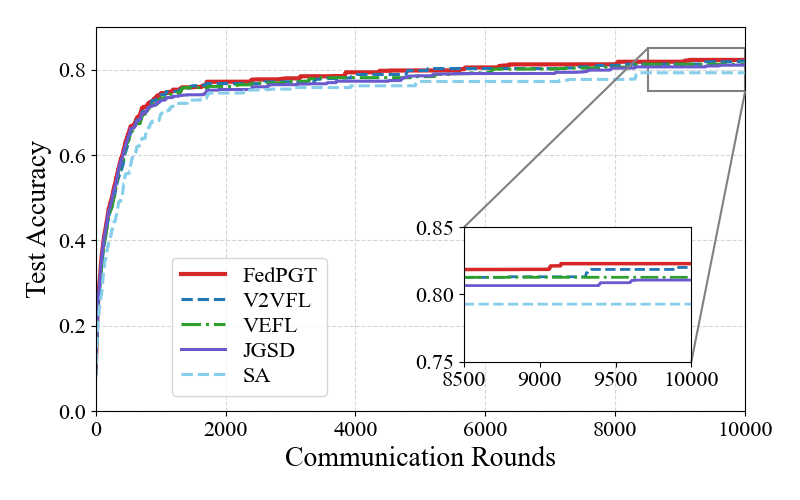}
    \caption{Convergence performance when the total bandwidth is $30$ MHz.}
    \label{bandwidth30}
  \end{minipage}
  \hfill
  \begin{minipage}[t]{0.32\textwidth}
    \vspace{0pt}
    \centering
    \includegraphics[width=\textwidth]{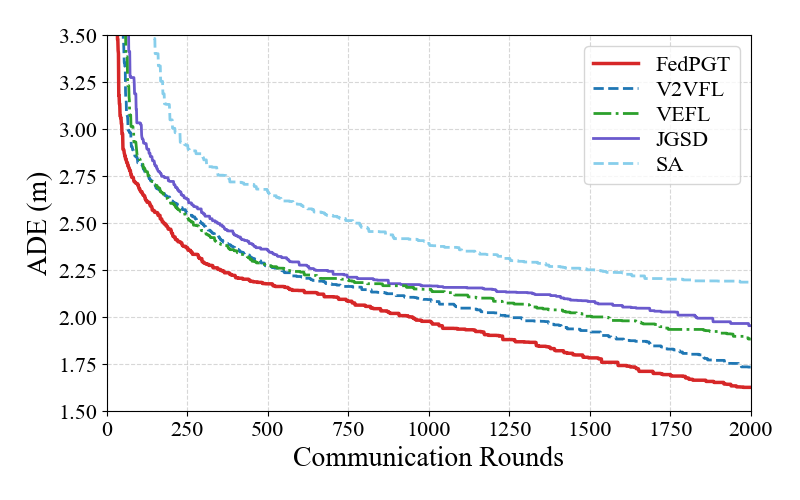}
    \caption{ADE for the trajectory prediction task when the total bandwidth is $20$ MHz and $v^{\max}=5$ m/s.}
    \label{ADE1}
  \end{minipage}
    \hfill
  \begin{minipage}[t]{0.32\textwidth}
    \vspace{0pt}
    \centering
    \includegraphics[width=\textwidth]{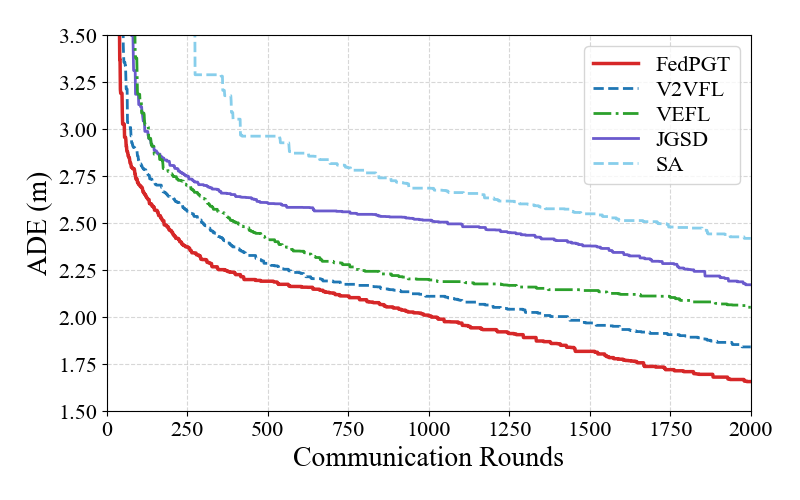}
    \caption{ADE for the trajectory prediction task when the total bandwidth is $20$ MHz and $v^{\max}=25$ m/s.}
    \label{ADE2}
  \end{minipage}
  \begin{minipage}[t]{0.32\textwidth}
    \vspace{0pt} 
    \centering
    \includegraphics[width=\textwidth]{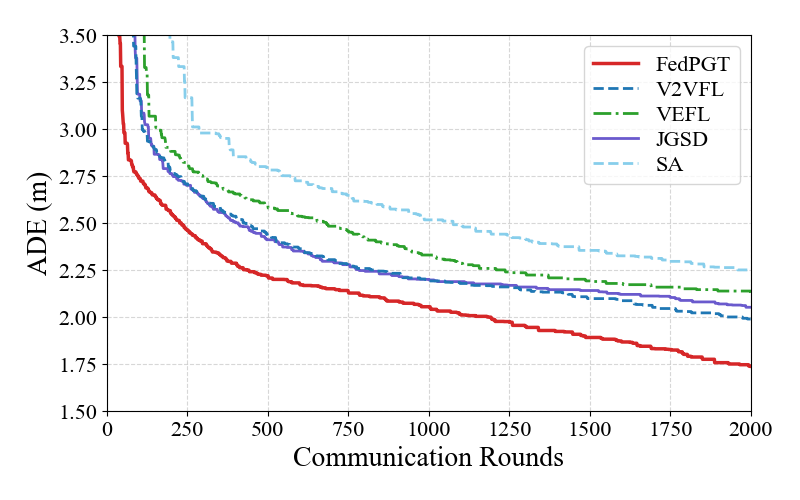}
    \caption{ADE for the trajectory prediction task when the total bandwidth is $10$ MHz and $v^{\max}=5$ m/s.}
    \label{ADE3}
  \end{minipage}
  \hfill
\end{figure*}

\subsection{Performance under Different Numbers of Vehicles}

We evaluate the performance of different schemes under varying average numbers of vehicles within the RSU coverage, as shown in Fig.~\ref{number}, while Figs.~\ref{number5} and \ref{number25} present the convergence performance with average numbers of vehicles of 5 and 25, respectively. As the number of vehicles increases, FedPGT and JGSD achieve steady performance improvement, while V2VFL, VEFL, and SA exhibit diminishing gains and gradually saturate. When the number of vehicles is small, the total bandwidth is sufficient to support most transmissions, and a large fraction of vehicles can successfully upload their updates. As a result, increasing the number of participants improves the convergence performance across all schemes. 

As the number of vehicles increases, the total communication demand gradually exceeds the available resources. In this case, schemes requiring complete update transmissions (V2VFL, VEFL, and SA) cannot efficiently exploit the increasing number of vehicles, since only a limited number of updates can be successfully aggregated within each communication round. In contrast, FedPGT and JGSD continue to achieve stable performance gains as the number of vehicles grows, since partially transmitted gradient entries can still contribute to model aggregation under limited communication resources. This also verifies Theorem~1: collecting dominant gradient entries from more vehicles is more beneficial than collecting more complete updates from fewer vehicles.

\subsection{Performance under Different Total Bandwidth}

We evaluate performance under varying total bandwidth. Fig.~\ref{bandwidth} shows the final accuracy, while Figs.~\ref{bandwidth10} and \ref{bandwidth30} present the convergence performance under 10 MHz and 30 MHz bandwidth settings, respectively. When the bandwidth is sufficient (30 MHz), all schemes achieve comparable performance, since most vehicles can successfully complete their update transmissions. As the total bandwidth decreases from 30 MHz to 10 MHz, the overall performance degrades due to limited communication resources.

In the low-bandwidth regime, schemes requiring complete update transmissions (V2VFL, VEFL, and SA) experience significant performance degradation, since limited bandwidth restricts the number of updates that can be successfully aggregated within each communication round. In contrast, FedPGT remains the most robust under limited bandwidth, as progressive gradient transmission enables dominant gradient entries from more vehicles to still contribute to model aggregation.

\subsection{Evaluation on Argoverse Trajectory Prediction Dataset}

We further evaluate the performance of different schemes on the Argoverse trajectory prediction dataset. The results are shown in Figs.~\ref{ADE1}--\ref{ADE3}, where the average displacement error (ADE) is used as the evaluation metric. When the vehicle speed is low ($v^{\text{max}}=5$ m/s) and the bandwidth is sufficient (20 MHz), all schemes achieve comparable performance, with FedPGT achieving the lowest ADE. As the system operates under more challenging conditions, such as limited bandwidth (10 MHz) or high mobility ($v^{\text{max}}=25$ m/s), the performance gap between FedPGT and the baselines becomes more significant. In these cases, FedPGT exhibits a clear advantage, demonstrating its robustness to both communication constraints and mobility.  These results further verify that the advantages of FedPGT extend beyond image classification tasks to more complex spatio-temporal prediction problems.

\section{Conclusion}

In this paper, we proposed FedPGT, a progressive gradient transmission scheme for vehicular federated learning over time-varying wireless channels. We first established a convergence bound that reveals the diminishing-return effect of transmitted gradient entries, which motivates the optimization of progressive transmission according to marginal learning utility. To address the cumulatively coupled transmission objective, we developed an additively separable surrogate reformulation and introduced the PGT and energy queues to track transmission consistency and energy-budget deviation. This enables a Lyapunov-based tradeoff between learning utility and energy cost, allowing FedPGT to jointly adapt RB assignment and power allocation without future system information. Simulation results on CIFAR-10 and Argoverse demonstrated that FedPGT achieves more robust learning performance than existing vehicular federated learning schemes under high mobility and limited communication resources.

Future work will investigate the integration of semantic-aware transmission and more advanced gradient compression techniques into the proposed FedPGT framework to further reduce communication overhead while preserving learning performance. We will also extend the framework to more practical vehicular federated learning scenarios, such as asynchronous model aggregation under dynamic connectivity.

\bibliographystyle{./bibliography/IEEEtran}
\bibliography{./bibliography/IEEEabrv,./bibliography/IEEEexample}

\begin{IEEEbiography}[{\includegraphics[width=1in,height=1.25in,clip,keepaspectratio]{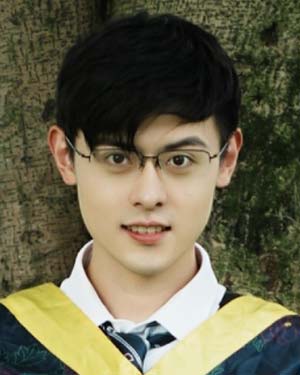}}]{Jintao Yan}(S'23)
received the B.S. degree from the University of Electronic Science and Technology of China (UESTC), Chengdu, China, in 2022. He is currently working toward the Ph.D. degree with the Network Integration for Ubiquitous Linkage and Broadband Laboratory (Niulab), Department of Electronic Engineering, Tsinghua University. His research interests include edge intelligence, vehicular networks, federated learning and optimization theory.
\end{IEEEbiography}

\begin{IEEEbiography}[{\includegraphics[width=1in,height=1.25in,clip,keepaspectratio]{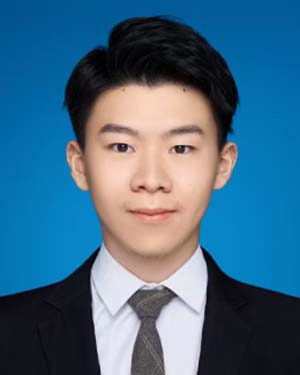}}]{Tan Chen}(S'24)
received the B.S. degree from the Department of Electronic Engineering, Tsinghua University, Beijing, China, in 2021. He is currently working toward the Ph.D. degree with the Network Integration for Ubiquitous Linkage and Broadband Laboratory (Niulab), Department of Electronic Engineering, Tsinghua University. His research interests include federated learning, vehicular networks and edge intelligence.
\end{IEEEbiography}

\begin{IEEEbiography}[{\includegraphics[width=1in,height=1.25in,clip,keepaspectratio]{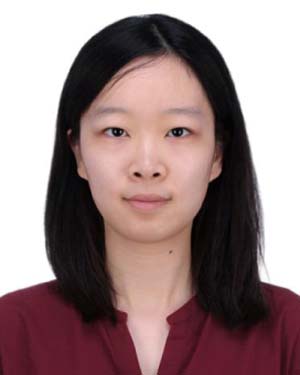}}]{Yuxuan Sun}(S'18-M'20)
received the Ph.D. degree in electronic engineering from Tsinghua University, Beijing, China, in 2020. From 2018 to 2019, she was a Visiting Student at the Department of Electrical and Electronic Engineering, Imperial College London, UK. From 2020 to 2022, she was a Post-Doctoral Researcher at the Department of Electronic Engineering, Tsinghua University, and a Visiting Researcher at Imperial College London. Currently, she is an Associate Professor at the School of Electronic and Information Engineering, Beijing Jiaotong University, Beijing, China. She is the receipt of the Young Elite Scientists Sponsorship Program by CAST.  She served as the Assistant to the Editor-in-Chief of IEEE \textsc{Transactions on Green Communications and Networking} from 2020 to 2022. She has been the Secretary of IEEE ComSoC Emerging Technologies Standing Committee since 2022. Her research interests lie in the areas of edge computing, edge intelligence, and task-oriented communications. 
\end{IEEEbiography}

\begin{IEEEbiography}[{\includegraphics[width=1in,height=1.25in,clip,keepaspectratio]{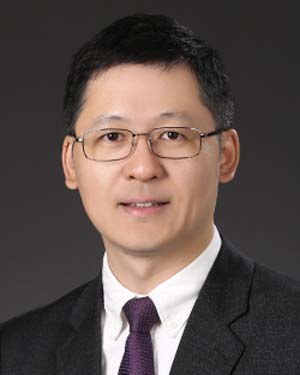}}]{Sheng Zhou} (S'06-M'12-SM'24)
received the B.E. and Ph.D. degrees in electronic engineering from Tsinghua University, Beijing, China, in 2005 and 2011, respectively. In 2010, he was a Visiting Student with the Wireless System Lab, Department of Electrical Engineering, Stanford University, Stanford, CA, USA. From 2014 to 2015, he was a Visiting Researcher with the Central Research Lab, Hitachi Ltd., Japan. He is currently an Associate Professor with the Department of Electronic Engineering, Tsinghua University. His research interests include cross-layer design for multiple antenna systems, mobile edge computing, vehicular networks, and green wireless communications. He received the IEEE ComSoc Asia–Pacific Board Outstanding Young Researcher Award in 2017, and IEEE ComSoc Wireless Communications Technical Committee Outstanding Young Researcher Award in 2020.
\end{IEEEbiography}

\begin{IEEEbiography}[{\includegraphics[width=1in,height=1.25in,clip,keepaspectratio]{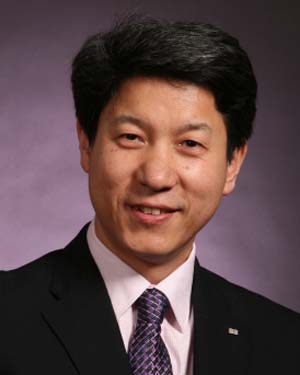}}]{Zhisheng Niu}(M'98-SM'99-F'12)
graduated from Beijing Jiaotong University, China, in 1985, and got his M.E. and D.E. degrees from Toyohashi University of Technology, Japan, in 1989 and 1992, respectively.  From 1992 to 1994, he worked for Fujitsu Laboratories Ltd., Japan, and in 1994 joined with Tsinghua University, Beijing, China, where he is now a Professor at the Department of Electronic Engineering. His major research interests include queueing theory, traffic engineering, mobile Internet, radio resource management of wireless networks, and green communication and networks.
 
Dr. Niu has been serving IEEE Communications Society since 2000 as Chair of Beijing Chapter (2000-2008), Director of Asia-Pacific Board (2008-2009), Director for Conference Publications (2010-2011), Chair of Emerging Technologies Committee (2014-2015), Director for Online Contents (2018-2019) and currently the Editor-in-Chief of IEEE \textsc{Transactions on Green Communications and Networking}. He received the Best Paper Award of Asia-Pacific Board in 2013, Distinguished Technical Achievement Recognition Award of Green Communications and Computing Technical Committee in 2018, and Harold Sobol Award for Exemplary Service to Meetings \& Conferences in 2019, all from the IEEE Communications Society. He was selected as a distinguished lecturer of IEEE Communications Society (2012-2015) as well as IEEE Vehicular Technologies Society (2014-2018). He is a fellow of both IEEE and IEICE.

\end{IEEEbiography}

\newpage

\appendices

\section{Proof of Lemma 1}
\label{Lemma1}
Ranking the elements of $\boldsymbol{g}^{(r)}_n$ in descending order, where $|g_{n,1}^{(r)}|\geq |g_{n,2}^{(r)}| \geq...\geq |g_{n,I}^{(r)}|$, we have
\begin{align}&\|\boldsymbol{g}_{n}^{(r)}-\Psi_n^{(r)}(\boldsymbol{g}_{n}^{(r)})\|^{2}\notag\\&=\sum_{i=k_n^{(r)}+1}^I(g_{n,i}^{(r)})^2\leq\sum_{i=k_n^{(r)}+1}^I (C_n^{(r)})^2 i^{-2\alpha_n^{(r)}}\notag\\&\leq (C_n^{(r)})^2 \left(k_n^{(r)}+1\right)^{-2\alpha_n^{(r)}}+\sum_{i=k_n^{(r)}+2}^\infty (C_n^{(r)})^2i^{-2\alpha_n^{(r)}} \notag\\ &\leq  (C_n^{(r)})^2 \left(k_n^{(r)}+1\right)^{-2\alpha_n^{(r)}} + (C_n^{(r)})^2\int_{k_n^{(r)}+1}^\infty x^{-2\alpha_n^{(r)}}dx\notag\\&=(C_n^{(r)})^2 \left((k_n^{(r)}+1)^{-2\alpha_n^{(r)}}+\frac{(k_n^{(r)}+1)^{-(2\alpha_n^{(r)}-1)}}{2\alpha_n^{(r)}-1}\right).\notag\\&\leq  \frac{2(C_n^{(r)})^2\alpha_n^{(r)}(k_n^{(r)}+1)^{-(2\alpha_n^{(r)}-1)}}{2\alpha_n^{(r)}-1}\notag\end{align}
This completes the proof.

\section{Proof of Theorem 1}
\label{Theorem1}

According to Assumption 4 and definition (\ref{global}), the global loss function is also $L$-smooth. There is:
\begin{align}F&(\boldsymbol {w}^{(r)}) - F(\boldsymbol {w}^{(r-1)}) \label{conv1} \\
&\leq \left < { \nabla F(\boldsymbol {w}^{(r-1)}), \boldsymbol {w}^{(r)}-\boldsymbol {w}^{(r-1)} }\right >  + \frac {L}{2} \left \Vert{ \boldsymbol {w}^{(r)}-\boldsymbol {w}^{(r-1)} }\right \Vert ^{2}.\notag
\end{align}
For the term $\left < { \nabla F(\boldsymbol {w}^{(r-1)}), \boldsymbol {w}_{r+1}-\boldsymbol {w}^{(r)} }\right >$, we have 
\begin{align}
& \left < { \nabla F(\boldsymbol {w}^{(r-1)}), \boldsymbol {w}^{(r)}-\boldsymbol {w}^{(r-1)} }\right > \notag\\ 
&=  -\eta  \big < { \nabla F(\boldsymbol {w}^{(r-1)}), \frac{1}{N^{(r)}}\sum_{n\in \mathcal{N}^{(r)}} \Psi_n^{(r)}(\boldsymbol{g}_n^{(r)})} \big >\notag\\
&= -\frac{\eta}{2}\left \Vert{\nabla F(\boldsymbol {w}^{(r-1)})}\right \Vert^2 - \frac{\eta}{2}   \left \Vert{ \frac{\sum_{n\in \mathcal{N}^{(r)}}\Psi_n^{(r)}(\boldsymbol{g}_n^{(r)})}{N^{(r)}}}\right \Vert ^{2}\notag \\
&  + \frac{\eta}{2}\left \Vert{\nabla F(\boldsymbol {w}^{(r-1)}) -\frac{1}{N^{(r)}}\sum_{n\in \mathcal{N}^{(r)}} \Psi_n^{(r)}(\boldsymbol{g}_n^{(r)}) }\right \Vert^2 \notag \\
&= -\frac{\eta}{2}\left \Vert{\nabla F(\boldsymbol {w}^{(r-1)})}\right \Vert^2 - \frac{\eta}{2} \left \Vert{ \frac{\sum_{n\in \mathcal{N}^{(r)}}\Psi_n^{(r)}(\boldsymbol{g}_n^{(r)})}{N^{(r)}}}\right \Vert ^{2}\label{term1}
\\
&+ \eta\left \Vert{\nabla F(\boldsymbol {w}^{(r-1)}) -\frac{\sum_{n\in \mathcal{N}^{(r)}}\boldsymbol{g}_n^{(r)}}{N^{(r)}} }\right \Vert^2 \notag \\
& + \eta \left \Vert{\frac{\sum_{n\in \mathcal{N}^{(r)}}\left(\boldsymbol{g}_n^{(r)}- \Psi_n^{(r)}(\boldsymbol{g}_n^{(r)})\right)}{N^{(r)}} }\right \Vert^2 \notag\\
&\leq -\frac{\eta}{2}\left \Vert{\nabla F(\boldsymbol {w}^{(r-1)})}\right \Vert^2 -  \frac{\eta}{2} \left \Vert{ \frac{\sum_{n\in \mathcal{N}^{(r)}}\Psi_n^{(r)}(\boldsymbol{g}_n^{(r)})}{N^{(r)}}}\right \Vert ^{2}\notag\\&+ \frac{\eta (\sigma^2+\delta^2)}{N^{(r)}} +  \sum_{n\in \mathcal{N}^{(r)}} \frac{2\eta(C_n^{(r)})^2\alpha_n^{(r)} (k_n^{(r)}+1)^{-(2\alpha_n^{(r)}-1)}}{N^{(r)}(2\alpha_n^{(r)}-1)}.\notag
\end{align}
For the term $\left \Vert{ \boldsymbol {w}^{(r)}-\boldsymbol {w}^{(r-1)} }\right \Vert^2$, we have 
\begin{equation}
\begin{aligned}
& \frac {L}{2} \left \Vert{ \boldsymbol {w}^{(r)}-\boldsymbol {w}^{(r-1)} }\right \Vert ^{2} = \frac {\eta^2 L}{2}   \left \Vert{ \frac{\sum_{n\in \mathcal{N}^{(r)}}\Psi_n^{(r)}(\boldsymbol{g}_n^{(r)})}{N^{(r)}}}\right \Vert ^{2}.\label{term2}
\end{aligned}
\end{equation}
Setting $\eta \leq \frac{1}{L}$, taking the expectation over stochastic data sampling on (\ref{conv1}) and plugging (\ref{term1}) and (\ref{term2}), we have
\begin{equation}
\begin{aligned}
&\frac{\eta}{2}\left \Vert{\nabla F(\boldsymbol {w}^{(r-1)})}\right \Vert^2   \leq  \mathbb{E}[F(\boldsymbol{w}^{(r-1)})] - \mathbb{E}[F(\boldsymbol{w}^{(r)})]\\ 
&+ \frac{\eta (\sigma^2+\delta^2)}{N^{(r)}} + \sum_{n\in \mathcal{N}^{(r)}} \frac{2\eta(C_n^{(r)})^2\alpha_n^{(r)} (k_n^{(r)}+1)^{-(2\alpha_n^{(r)}-1)}}{N^{(r)}(2\alpha_n^{(r)}-1)}.\notag
\end{aligned}
\end{equation}
Taking a telescopic sum from $r = 1$ to $r = R$, we get
\begin{equation}
\begin{aligned}
&\frac{1}{R}\sum_{r=0}^{R-1} \mathbb{E}\left\|\nabla F(\boldsymbol {w}^{(r)})\right\|^2 \\& \leq \frac{2\left(F(\boldsymbol{w}^{(0)})-\mathbb{E}[F(\boldsymbol{w}^{(R)})]\right)}{\eta R}+\frac{1}{R}\sum_{r=1}^{R}\frac{2(\sigma^2+\delta^2)}{N^{(r)}}\\
& +\frac{1}{R}\sum_{r=1}^{R} \sum_{n\in \mathcal{N}^{(r)}} \frac{4(C_n^{(r)})^2\alpha_n^{(r)} (k_n^{(r)}+1)^{-(2\alpha_n^{(r)}-1)}}{N^{(r)}(2\alpha_n^{(r)}-1)}.  \notag
\end{aligned}
\end{equation}
This completes the proof.

\section{Proof of Theorem 2}
\label{Theorem2}
We define a quadratic Lyapunov function as $$L_n(t) \triangleq\frac{1}{2} \sum_{n\in \mathcal{N}^{(r)}}(q_{n}(t))^2+\frac{1}{2} \sum_{n\in \mathcal{N}^{(r)}}(\zeta_{n}(t))^2,$$
and the Lyapunov drift of a single round as
\begin{align}
    &\Delta_n(t)\triangleq L_n(t+1)-L_n(t) \notag\\&= \frac{1}{2} \left( q_{n}(t+1))^2 - (q_{n}(t))^2 + (\zeta_{n}(t+1))^2 - (\zeta_{n}(t))^2 \right)\notag\\ 
    &\leq\frac{1}{2} \left( \left(q_{n}(t)+\epsilon_{n}(t)\right)^2-(q_{n}(t))^2 \right) \label{lypdrift} \\& + \frac{1}{2} \left( \left(\zeta_{n}(t)+\kappa_n(t)-\gamma_n(t)\right)^2-(\zeta_{n}(t))^2 \right)\notag\\& \leq \Phi+ q_{n}(t) \epsilon_{n}(t) +  \zeta_{n}(t)\left(\kappa_n(t)-\gamma_n(t)\right),\notag
\end{align}
where $\epsilon_{n}(t) \triangleq  e_{n}(t)- \frac{E^{\text{cons}}_n-\xi_n^{(r)}}{T}$, $\phi_n \triangleq \max_t\{|\epsilon_{n}(t)|\}$, $\varphi_n\triangleq\max_t\{|\kappa_n(t)-\gamma_n(t)|\}$ and $\Phi \triangleq\max_n\{(\phi_n)^2+(\varphi_n)^2\}$. We define the overall drift as $\Delta_n \triangleq  \Delta_n(T+1)-\Delta_n(1)$, which is bounded by
\begin{equation}
\begin{aligned}
\Delta_n \leq T \Phi+ \sum_{t\in\mathcal{T}^{(r)}} \left( q_{n}(t) \epsilon_{n}(t) +  \zeta_{n}(t)\left(\kappa_n(t)-\gamma_n(t)\right)\right).\notag
\end{aligned}
\end{equation}
Based on the definition of $q_{n}(t)$, we have $q_{n}(t+1) - q_{n}(t) \leq \phi_n, \forall n \in \mathcal{N},r \in \mathcal{R}$. Therefore, for any $\epsilon_{n}(t)$, we have
\begin{equation}
\begin{aligned}
q_{n}(t) \epsilon_{n}(t) \leq (q_{n}(t)-q_{n}(1)) \phi_n \leq (t-1) (\phi_n)^2.\notag
\end{aligned}
\end{equation}
Similarly, there is 
\begin{equation}
\begin{aligned}
\zeta_{n}(t)(\kappa_n(t)-\gamma_n(t))  \leq (t-1) (\varphi_n)^2.\notag
\end{aligned}
\end{equation}
Therefore, we have
\begin{equation}
\begin{aligned}
\Delta_n \leq \sum_{t\in\mathcal{T}^{(r)}} (t-1) \Phi =T^2 \Phi.\notag
\end{aligned}
\end{equation}
Based on this, we have
\begin{equation}
\begin{aligned}&\left|\sum_{t\in\mathcal{T}^{(r)}}\kappa_n(t)-\gamma_n(t)\right|=\left|\sum_{t\in\mathcal{T}^{(r)}}\zeta_n(t+1)-\zeta_n(t)\right|\\&=|\zeta_n(T+1)|\leq\sqrt{2\Delta_n}\leq T\sqrt{2 \Phi} \end{aligned}
\end{equation}
For energy consumption, we have
\begin{equation}
\begin{aligned}
&\sum_{t\in\mathcal{T}^{(r)}} \left( e_{n}^\dagger(t)- \frac{E^{\text{cons}}_n-\xi_n^{(r)}}{T} \right) \notag\\ &\leq \sum_{t\in\mathcal{T}^{(r)}} q_{n}{(t+1)} - q_{n}(t)\leq T\sqrt{2 \Phi}.\notag
\end{aligned}
\end{equation}
Adding $\frac{V(C_n^{(r)})^2\alpha_n^{(r)}\left(\gamma_n^\dagger(t)+\frac{1}{T}\right)^{-(2\alpha_n^{(r)}-1)}}{2 \alpha_n^{(r)}-1}$ on both sides of (\ref{lypdrift}), the single-round drift-plus-penalty function is bounded by
\begin{equation}
\begin{aligned}
&\Delta_n(t) + \frac{V(C_n^{(r)})^2\alpha_n^{(r)}\left(\gamma_n^{\dagger}(t)+\frac{1}{T}\right)^{-(2\alpha_n^{(r)}-1)}}{2 \alpha_n^{(r)}-1}
\\& \leq \Phi+  q_{n}(t) \epsilon_{n}^\dagger(t) +  \zeta_{n}(t)\left(\kappa_n^\dagger(t)-\gamma_n^\dagger(t)\right)\\&
+\frac{V(C_n^{(r)})^2\alpha_n^{(r)}\left(\gamma_n^{\dagger}(t)+\frac{1}{T}\right)^{-(2\alpha_n^{(r)}-1)}}{2 \alpha_n^{(r)}-1}\\&
\leq  \Phi+q_{n}(t) \epsilon_{n}^*(t) +  \zeta_{n}(t)\left(\kappa_{n}^*(t)-\frac{k_n^{(r)*}}{T}\right)\\&+\frac{V(C_n^{(r)})^2\alpha_n^{(r)}}{2 \alpha_n^{(r)}-1}\left(\frac{k_n^{(r)*}+1}{T}\right)^{-(2\alpha_n^{(r)}-1)}.
\end{aligned}
\end{equation}
Taking a telescopic sum for both sides, we have
\begin{equation}
\begin{aligned}
&\Delta_n+ \sum_{t\in\mathcal{T}^{(r)}} \frac{V(C_n^{(r)})^2\alpha_n^{(r)}\left(\gamma_n^{\dagger}(t)+\frac{1}{T}\right)^{-(2\alpha_n^{(r)}-1)}}{2 \alpha_n^{(r)}-1} \\& \leq T^2 \Phi + 
\frac{ T^{2\alpha_n^{(r)}} V(C_n^{(r)})^2\alpha_n^{(r)}(k_n^{(r)*}+1)^{-(2\alpha_n^{(r)}-1)}}{2 \alpha_n^{(r)}-1}.\label{tele}
\end{aligned}
\end{equation}
For any positive sequence $\{\kappa_n(t)\}_{t\in\mathcal{T}^{(r)}}$, the function 
$f(x)=x^{-(2\alpha_n^{(r)}-1)}$ is convex and monotonically decreasing over $x>0$. 
By Jensen's inequality, we have
\begin{equation}
\begin{aligned}&T^{2\alpha_n^{(r)}}\bigg(\sum_{t\in\mathcal{T}^{(r)}}\gamma_n^\dagger(t)+1\bigg)^{-(2\alpha_n^{(r)}-1)} \\ &\leq \sum_{t\in\mathcal{T}^{(r)}}\left(\gamma_n^\dagger(t)+\frac{1}{T}\right)^{-(2\alpha_n^{(r)}-1)}\label{util1}\end{aligned}
\end{equation}
Also, since the function $g(x) = (x+1)^{-(2\alpha_n^{(r)}-1)}$ is continuously differentiable and Lipschitz continuous on $x \geq 0$, and its derivative satisfies
\[
\sup_{x \geq 0} \left| g'(x) \right|
= \sup_{x \geq 0} (2\alpha_n^{(r)} - 1)(x+1)^{-2\alpha_n^{(r)}}
\leq 2\alpha_n^{(r)} - 1,
\]
it follows from the mean value theorem that, for any $A,B \geq 0$,
\[
g(x) \leq g(y) + (2\alpha_n^{(r)} - 1) |x - y|.
\]
Therefore, we have
\begin{align}
&\bigg(\sum_{t \in \mathcal{T}^{(r)}} \kappa_n^\dagger(t) + 1 \bigg)^{-(2\alpha_n^{(r)} - 1)}\leq \bigg(\sum_{t \in \mathcal{T}^{(r)}} \gamma_n^\dagger(t) + 1 \bigg)^{-(2\alpha_n^{(r)} - 1)}\notag \\
&
+ (2\alpha_n^{(r)} - 1)\left| \sum_{t \in \mathcal{T}^{(r)}} \kappa_n^\dagger(t)-\gamma_n^\dagger(t) \right| \label{util2}\\
&\leq \bigg(\sum_{t \in \mathcal{T}^{(r)}} \gamma_n^\dagger(t) + 1 \bigg)^{-(2\alpha_n^{(r)} - 1)}
+ (2\alpha_n^{(r)} - 1) T \sqrt{2\Phi}.\notag
\end{align}
Substituting (\ref{util1}) and (\ref{util2}) into (\ref{tele}), we have
\begin{equation}
\begin{aligned}
&\sum_{n\in\mathcal{N}^{(r)}}\frac{(C_n^{(r)})^2\alpha_n^{(r)}}{2 \alpha_n^{(r)}-1}\bigg(\sum_{t\in\mathcal{T}^{(r)}}\kappa_n^\dagger(t)+1\bigg)^{-(2\alpha_n^{(r)}-1)} \\& \leq \sum_{n\in \mathcal{N}^{(r)}} \frac{(C_n^{(r)})^2\alpha_n^{(r)}(k_n^{(r)*}+1)^{-(2\alpha_n^{(r)}-1)}}{2 \alpha_n^{(r)}-1}\\&+\sum_{n\in \mathcal{N}^{(r)}}\frac{T^{2-2\alpha_n^{(r)}} \Phi}{V}+(C_n^{(r)})^2\alpha_n^{(r)} T \sqrt{2\Phi}.\notag
\end{aligned}
\end{equation}
Theorem 2 is proved.

\section{Proof of Proposition 1}
\label{Proposition1}
The Lagrangian of $P3$ is given by:
\begin{align}
&\mathcal{L} = \sum_{n\in\mathcal{N}^{(r)}} \sum_{z\in\mathcal{Z}}  \frac{\zeta_n(t)\tau \beta s_{n,z}(t) \log_2 (1+  \Gamma_{n,z}(t))}{b+\lceil \log_2 I \rceil} \notag\\
&+\sum_{n\in\mathcal{N}^{(r)}} \sum_{z\in\mathcal{Z}} \tau q_{n}(t) s_{n,z}(t) p_{n,z}(t) \\ &+ \sum_{z\in\mathcal{Z}}\lambda_z  (\sum_{n\in\mathcal{N}^{(r)}}  s_{n,z}(t) - 1 ) - \sum_{n\in\mathcal{N}^{(r)}} \sum_{z\in\mathcal{Z}} \varsigma_{n,z} s_{n,z}(t)\notag\\&+
\sum_{n\in\mathcal{N}^{(r)}} \nu_n (\sum_{z\in\mathcal{Z}}  p_{n,z}(t) - p^{\text{max}}_n  ) - \sum_{n\in\mathcal{N}^{(r)}} \sum_{z\in\mathcal{Z}} \chi_{n,z} p_{n,z}(t).\notag
\end{align}

Then the KKT condition is given by:
\begin{subequations}
\begin{align}
& \frac{\zeta_n(t)\tau \beta \log_2 (1+  \Gamma_{n,z}(t))}{b+\lceil \log_2 I \rceil}  + \tau q_{n}(t) p_{n,z}(t) \notag \\& +\lambda_{z}-\varsigma_{n,z}= 0, \quad \forall n\in\mathcal{N}^{(r)}, \forall z\in \mathcal{Z}, \label{kkt1} \\
& \frac{\zeta_n(t)\tau \beta s_{n,z}(t)\frac{\Vert \boldsymbol{h}_{n,z}(t) \big\Vert^2}{\beta N_0}}{\Big(1+ \frac{p_{n,z}(t) \big\Vert \boldsymbol{h}_{n,z}(t) \big\Vert^2}{\beta N_0}\Big)\ln2} + \tau q_{n}(t) s_{n,z}(t) \notag \\& +\nu_{n} - \chi_{n,z} = 0,  \quad \forall n\in\mathcal{N}^{(r)}, \forall z\in \mathcal{Z}, \label{kkt2}\\
& \lambda_z (\sum_{n\in\mathcal{N}^{(r)}}  s_{n,z}(t) - 1) = 0, \quad  \forall z\in \mathcal{Z}, \label{kkt3}\\
& \nu_n (\sum_{z\in\mathcal{Z}_{n}(t)}  p_{n,z}(t) - p^{\text{max}}_n) = 0,\quad \forall n\in\mathcal{N}^{(r)},\label{kkt4}\\
& \varsigma_{n,z} s_{n,z}(t)=0, \quad \forall n\in\mathcal{N}^{(r)}, \forall z\in \mathcal{Z},\label{kkt5}\\
& \chi_{n,z} p_{n,z}(t)=0, \quad \forall n\in\mathcal{N}^{(r)}, \forall z\in \mathcal{Z},\label{kkt6}\\
& \lambda_{n,z}, \varsigma_{n,z}, \nu_n, \chi_{n,z}\geq 0, \quad \forall n\in\mathcal{N}^{(r)}, \forall z\in \mathcal{Z}, \label{kkt7}
\end{align}
\end{subequations}
where $\lambda_z$, $\varsigma_{n,z}$, $\nu_n$ and $\chi_{n,z}$ are non-negative Lagrange multipliers. Combining (\ref{kkt1}) and (\ref{kkt5}), for every $n\in \mathcal{N}^{(r)}$ and $z \in \mathcal{Z}$, we have
\begin{equation}
s_{n,z}(t) \left( \frac{\zeta_n(t)\tau \beta \log_2 (1+  \Gamma_{n,z}(t))}{b+\lceil \log_2 I \rceil}  + \tau q_{n}(t) p_{n,z}(t)  + \lambda_{z} \right)= 0. \notag 
\end{equation}
Since $\varsigma_{n,z}\geq0$, (\ref{kkt1}) implies that
\begin{equation}
\frac{\zeta_n(t)\tau \beta \log_2 (1+  \Gamma_{n,z}(t))}{b+\lceil \log_2 I \rceil}  + \tau q_{n}(t) p_{n,z}(t) \geq -\lambda_z. \label{p1upper} 
\end{equation}
If $s_{n,z}(t)$ is non-zero, the left-hand side of (\ref{p1upper}) reaches its lower bound, $\lambda_z$. Therefore, the proposition is proved.

\section{Proof of Proposition 3}
\label{Proposition3}

If $\zeta_n(t)\ge 0$, the objective function is nondecreasing with respect to each $p_{n,z}(t)$. Therefore, the optimal solution is
\begin{equation}
p_{n,z}(t)=0,\quad \forall z\in\mathcal Z_n(t).\notag
\end{equation}

If $\zeta_n(t)< 0$, we combine (\ref{kkt2}) and (\ref{kkt6}) to obtain
\begin{equation}
p_{n,z}(t) \left( \frac{\zeta_n(t)\tau \beta s_{n,z}(t)\frac{\Vert \boldsymbol{h}_{n,z}(t) \big\Vert^2}{\beta N_0}}{\Big(1+ \frac{p_{n,z}(t) \big\Vert \boldsymbol{h}_{n,z}(t) \big\Vert^2}{\beta N_0}\Big)\ln2} + \tau q_{n}(t) s_{n,z}(t) \notag +\nu_{n} \right)=0.
\end{equation}
Then, we have
$$p_{n,z}(t) = \max\left( \frac{-\zeta_n(t)\tau \beta}{\left(\tau q_{n}(t)+\nu_n\right)\ln2} - \frac{\beta N_0}{\big\Vert \boldsymbol{h}_{n,z}(t) \big\Vert^2},0\right).$$
if $\nu_n > 0$, the total power constraint is active, i.e.,
$$\sum_{z\in\mathcal{Z}_{n}^*(t)}\left(\frac{-\zeta_n(t)\tau \beta}{\left(\tau q_{n}(t)+\nu_n\right)\ln2} - \frac{\beta N_0}{\big\Vert \boldsymbol{h}_{n,z}(t) \big\Vert^2}\right) = p^{\text{max}}_n,$$
where $\mathcal{Z}_{n}^*(t)$ is the subset of RBs to which vehicle $n$ assigns positive power. Then, there is
$$\nu_n =\frac{-\vert \mathcal{Z}_{n}^*(t)\vert \zeta_n(t)\tau \beta}{\bigg(p^{\text{max}}_n+\sum_{z\in\mathcal{Z}_{n}^*(t)}\frac{\beta N_0}{\big\Vert \boldsymbol{h}_{n,z}(t) \big\Vert^2}\bigg)\ln2}-\tau q_{n}(t).$$
Substituting $\nu_n$ back gives
$$p_{n,z}(t) = \frac{p^{\text{max}}_n+\sum_{z'\in\mathcal{Z}_{n}^*(t)}\frac{\beta N_0}{\big\Vert \boldsymbol{h}_{n,z'}(t) \big\Vert^2}}{\vert \mathcal{Z}_{n}^*(t)\vert} - \frac{\beta N_0}{\big\Vert \boldsymbol{h}_{n,z}(t)\big\Vert^2}.$$
If $\nu_n = 0$, the optimal power allocation becomes
$$p_{n,z}(t) = \frac{-\zeta_n(t) \beta}{ q_{n}(t)\ln2} - \frac{\beta N_0}{\big\Vert \boldsymbol{h}_{n,z}(t) \big\Vert^2}.$$

Combining the above cases, when $\zeta_n(t)<0$ and $q_n(t)>0$, the optimal power allocation is given by
\begin{align}
p_{n,z}(t)
=&
\min \left(
\frac{
p^{\max}_n+
\sum_{z'\in\mathcal{Z}_{n}^*(t)}
\frac{\beta N_0}{\big\Vert \boldsymbol{h}_{n,z'}(t) \big\Vert^2}
}
{\vert \mathcal{Z}_{n}^*(t)\vert},
\frac{-\zeta_n(t)\beta}{q_n(t)\ln 2}
\right)
\notag\\
&-
\frac{\beta N_0}{\big\Vert \boldsymbol{h}_{n,z}(t)\big\Vert^2},
\quad
\forall z \in \mathcal{Z}_{n}^*(t).\notag
\end{align}

When $\zeta_n(t)<0$ and $q_n(t)=0$, the objective function is monotonically decreasing with respect to the achievable transmission rate. Therefore, the total power constraint is active, and the optimal power allocation reduces to
\begin{equation}
\begin{aligned}
p_{n,z}(t)
=&
\frac{
p_n^{\max}
+
\sum_{z'\in\mathcal Z_n^*(t)}
\frac{\beta N_0}{\|\boldsymbol h_{n,z'}(t)\|^2}
}
{|\mathcal Z_n^*(t)|}
-
\frac{\beta N_0}{\|\boldsymbol h_{n,z}(t)\|^2},
\\ &\forall z\in\mathcal Z_n^*(t).\notag
\end{aligned}
\end{equation}
Therefore, the proposition is proved.



\end{document}